\documentclass[11pt]{article}
\usepackage[preprint]{acl}
\usepackage{times}
\usepackage{latexsym}
\usepackage[T1]{fontenc}
\usepackage[utf8]{inputenc}
\usepackage{microtype}
\usepackage{inconsolata}
\usepackage{graphicx}
\usepackage{hyperref}
\usepackage{url}
\usepackage{graphicx}
\usepackage[table,xcdraw]{xcolor}
\usepackage{colortbl}
\usepackage{tabularx}
\usepackage{booktabs}
\usepackage{multirow}
\usepackage{wrapfig}
\usepackage{caption} 
\usepackage{enumitem}
\usepackage{amsmath}
\usepackage{array}
\usepackage{tikz}
\usepackage{listings}
\usepackage{xcolor}
\usepackage{ragged2e}
\usepackage{tcolorbox}
\usepackage{enumitem}
\usepackage{hyperref}
\newcolumntype{L}[1]{>{\RaggedRight\arraybackslash}p{#1}} 
\usepackage{graphicx}
\usepackage{subcaption}
\usepackage{caption}
\usepackage{float}

\usepackage[utf8]{inputenc} 
\usepackage[T1]{fontenc}    
\usepackage{hyperref}       
\usepackage{url}            
\usepackage{booktabs}       
\usepackage{amsfonts}       
\usepackage{nicefrac}       
\usepackage{microtype}      
\usepackage{xcolor}         
\usepackage[utf8]{inputenc}
\usepackage[T1]{fontenc}
\usepackage{ifthen}

\usepackage{hyperref}
\usepackage{url}
\usepackage{booktabs}
\usepackage{amsfonts}
\usepackage{nicefrac}
\usepackage{microtype}
\usepackage{graphicx}
\usepackage{array}
\usepackage{ifthen}      
\usepackage{tocloft}     
\let\oldaddcontentsline\addcontentsline
\renewcommand{\addcontentsline}[3]{}

\usepackage{makecell}
\usepackage{listings} 
\usepackage{xcolor}   
\definecolor{codegreen}{rgb}{0,0.6,0}
\definecolor{codegray}{rgb}{0.5,0.5,0.5}
\definecolor{codepurple}{rgb}{0.58,0,0.82}
\definecolor{backcolour}{rgb}{0.95,0.95,0.92}

\lstdefinestyle{mystyle}{
    backgroundcolor=\color{backcolour},
    basicstyle=\ttfamily\small\color{codegreen},
    stringstyle=\color{codegreen},
    keywordstyle=\color{codegreen},
    commentstyle=\color{codegreen},
    numberstyle=\tiny\color{codegray},
    breaklines=true,
    numbers=left,
    numbersep=5pt,
    showstringspaces=false,
    tabsize=2,
    lineskip=1.5pt,
    language=Python
}

\DeclareUnicodeCharacter{2019}{'}
\DeclareUnicodeCharacter{2014}{---}
\DeclareUnicodeCharacter{2013}{--}
\DeclareUnicodeCharacter{00B1}{\ensuremath{\pm}}
\DeclareUnicodeCharacter{2192}{\ensuremath{\to}}
\DeclareUnicodeCharacter{2264}{\ensuremath{\leq}}
\DeclareUnicodeCharacter{2265}{\ensuremath{\geq}}

\usepackage{listings}
\usepackage{listingsutf8}

\title{EngiBench: A Benchmark for Evaluating Large Language Models on Engineering Problem Solving}

\author{
\textbf{Xiyuan Zhou\textsuperscript{1}\thanks{Equal contribution.}},
\textbf{Xinlei Wang\textsuperscript{2,3}\footnotemark[1]},
\textbf{Yirui He\textsuperscript{4,5}},
\textbf{Yang Wu\textsuperscript{4}},
\textbf{Ruixi Zou\textsuperscript{4}},
\textbf{Yuheng Cheng\textsuperscript{4}},\\
\textbf{Yulu Xie\textsuperscript{6}},
\textbf{Wenxuan Liu\textsuperscript{1}},
\textbf{Huan Zhao\textsuperscript{7}},
\textbf{Yan Xu\textsuperscript{1}\thanks{Corresponding authors.}},
\textbf{Jinjin Gu\textsuperscript{3}\footnotemark[2]},
\textbf{Junhua Zhao\textsuperscript{4,8}\footnotemark[2]}\vspace{2mm}
\\
\textsuperscript{1}Nanyang Technological University,
\textsuperscript{2}The University of Sydney,\\
\textsuperscript{3}INSAIT, Sofia University ``St. Kliment Ohridski'',\\
\textsuperscript{4}The Chinese University of Hong Kong, Shenzhen,
\textsuperscript{5}Shenzhen Loop Area Institute,\\
\textsuperscript{6}The University of Hong Kong,
\textsuperscript{7}Hong Kong Polytechnic University,
\textsuperscript{8}AIRS
\\
\texttt{xiyuan002@e.ntu.edu.sg, \{xinlei.wang,jinjin.gu\}@insait.ai} \\
\texttt{xuyan@ntu.edu.sg, zhaojunhua@cuhk.edu.cn}
}

\begin{document}
\maketitle

\begin{abstract}\label{sec: abstract}
Large language models (LLMs) have shown strong performance on mathematical reasoning under well-defined conditions. However, real-world engineering problems involve uncertainty, context, and open-ended settings that extend beyond symbolic computation. Existing benchmarks largely focus on well-defined or abstract reasoning and therefore fail to capture these complexities. We introduce EngiBench, a hierarchical benchmark designed to evaluate LLMs on solving engineering problems. It spans three levels of increasing difficulty (foundational knowledge retrieval, contextual reasoning, and open-ended modeling) and covers diverse engineering subfields. To facilitate a deeper understanding of model performance, we systematically rewrite each problem into three controlled variants (perturbed, knowledge-enhanced, and math abstraction), enabling us to separately evaluate the model's robustness, domain-specific knowledge, and mathematical reasoning abilities. Experimental results show clear performance stratification across difficulty levels: model accuracy declines with task complexity, degrades under minor perturbations, and remains substantially below human performance on high-level engineering tasks. These findings reveal that current LLMs still lack the high-level reasoning needed for real-world engineering, highlighting the need for future models with deeper and more reliable problem-solving capabilities. Our source code and data are available at 
\url{https://github.com/AI4Engi/EngiBench}.
\end{abstract}

\section{Introduction}\label{sec: introduction}

Large language models (LLMs) have demonstrated promising capabilities in a range of mathematical reasoning tasks, from foundational skills such as basic computation and structured problem-solving \citep{cobbe2021training}, multi-step reasoning \citep{shao2024deepseekmath,wei2022chain}, to more complex applications like mathematical modeling \citep{guo2025deepseek} and the generation or verification of mathematical proofs \citep{yang2023leandojo,lin2025goedel,ren2025deepseek}.
However, just using mathematical reasoning is not enough for real-world applications. 
In practice, many applications arise not in abstract mathematical settings but in engineering contexts, where problems are grounded in physical systems and must handle uncertainty and real-world constraints.
These characteristics require not only mathematical computation, but also broader capabilities to understand engineering contexts and solve complex engineering problems.

\begin{figure*}[t]
\centering
\captionsetup{font={small}, skip=8pt}
\includegraphics[width=1.0\textwidth]{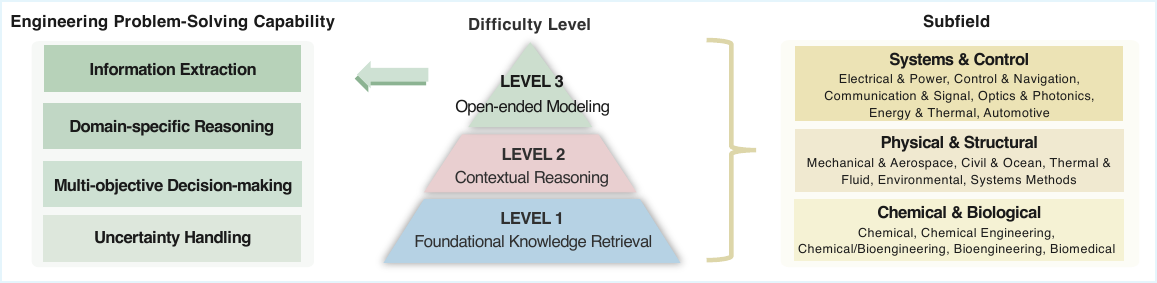}
\caption{
Task taxonomy of EngiBench organized by difficulty, capability, and subfield.
Problems are grouped into three difficulty levels, with Level 3 specifically designed to evaluate engineering problem-solving capabilities. 
All tasks are additionally categorised into three major engineering subfields.
}
\label{fig_0}
\end{figure*}

Engineering problems differ fundamentally from mathematical problems \citep{hendrycks2021measuring}. Rather than seeking single closed-form answers, engineering requires finding feasible solutions that balance objectives under real-world constraints \citep{dym2005engineering,zhou2026large}. For example, designing a drone system (Table~\ref{tab:Task hierarchy}) involves identifying operational requirements and managing trade-offs among range, payload, and energy limits. As shown in Figure~\ref{fig_0}, solving such problems requires more than recalling formulas or executing isolated calculations; it involves a sequence of interconnected cognitive steps, from understanding context and selecting appropriate assumptions to navigating trade-offs and addressing uncertainties. We refer to this broader set of competencies as the engineering problem-solving capability, consisting of four dimensions: \textit{information extraction, domain-specific reasoning, multi-objective decision-making, and uncertainty handling}.

Despite the broader requirements of real-world engineering tasks, most existing benchmarks focus narrowly on well-defined mathematical problems. Benchmarks such as 
GSM8K \citep{cobbe2021training}, MATH \citep{hendrycks2021measuring}, and Omni-MATH \citep{gao2025omnimath} primarily assess symbolic reasoning, calculation, and formal problem-solving under clean and well-defined conditions. Although some include basic engineering questions, they fail to capture the deeper reasoning required for real-world problem solving \citep{hendrycks2021measuring,wang2024mmlu,albalak2025big,du2025supergpqa}. A further limitation is that many benchmarks rely on public datasets without systematic rewriting, increasing the risk of pretraining overlap and inflated scores \citep{deng-etal-2024-investigating,huang2025math,sainz2023nlp}. For example, GSM1k re-creates GSM8k-style questions to reduce overlap and observes performance drops of up to 8\% \citep{zhang2024careful}. Without such safeguards, evaluations may reflect memorization rather than genuine reasoning, providing limited insight into an LLM’s ability to address realistic engineering tasks.

In this work, we introduce EngiBench, an evaluation framework designed not only to assess LLMs’ engineering problem-solving capabilities but also to diagnose where and why these capabilities fail. The benchmark spans multiple engineering subfields and structures tasks into three difficulty levels that reflect the progression from foundational knowledge retrieval to contextual reasoning and open-ended modeling. To support fine-grained diagnosis, each problem is provided in three controlled variants that separate robustness, domain knowledge, and mathematical reasoning. Evaluation centers on four capability dimensions essential to engineering problem solving: \textit{information extraction, domain-specific reasoning, multi-objective decision-making, and uncertainty handling}. For open-ended tasks, we further adopt rubric-based evaluation using expert-designed scoring criteria to ensure consistent and reliable assessment. Together, these components create a diverse, high-quality, and contamination-aware benchmark for evaluating LLMs’ engineering problem-solving capabilities.

Experiment results show clear stratification across difficulty levels, with higher-level tasks highlighting distinct capability gaps. The perturbed variant leads to performance drops, even in strong models, revealing that prior evaluations may overestimate true generalization. 
Most importantly, current LLMs perform poorly on Level 3 tasks involving open-ended, high-level engineering reasoning and remain far below human experts. These findings suggest that today’s LLMs are still far from reliably addressing real-world engineering problems, leaving substantial room for future improvement.

Our contributions can be summarized as follows: 
(1) We are among the first to systematically evaluate LLMs on real-world engineering problems; (2) We design a hierarchical benchmark with three difficulty levels and multiple problem variants, enabling fine-grained analysis of model reasoning capabilities and limitations; (3) Unlike prior benchmarks, our benchmark systematically evaluates LLM performance on open-ended engineering tasks; (4) We evaluate a broad set of mainstream LLMs, providing insights that can aid future model development and enhance engineering capabilities.

\begin{table*}[ht]
\centering
\includegraphics[width=1.0\textwidth]{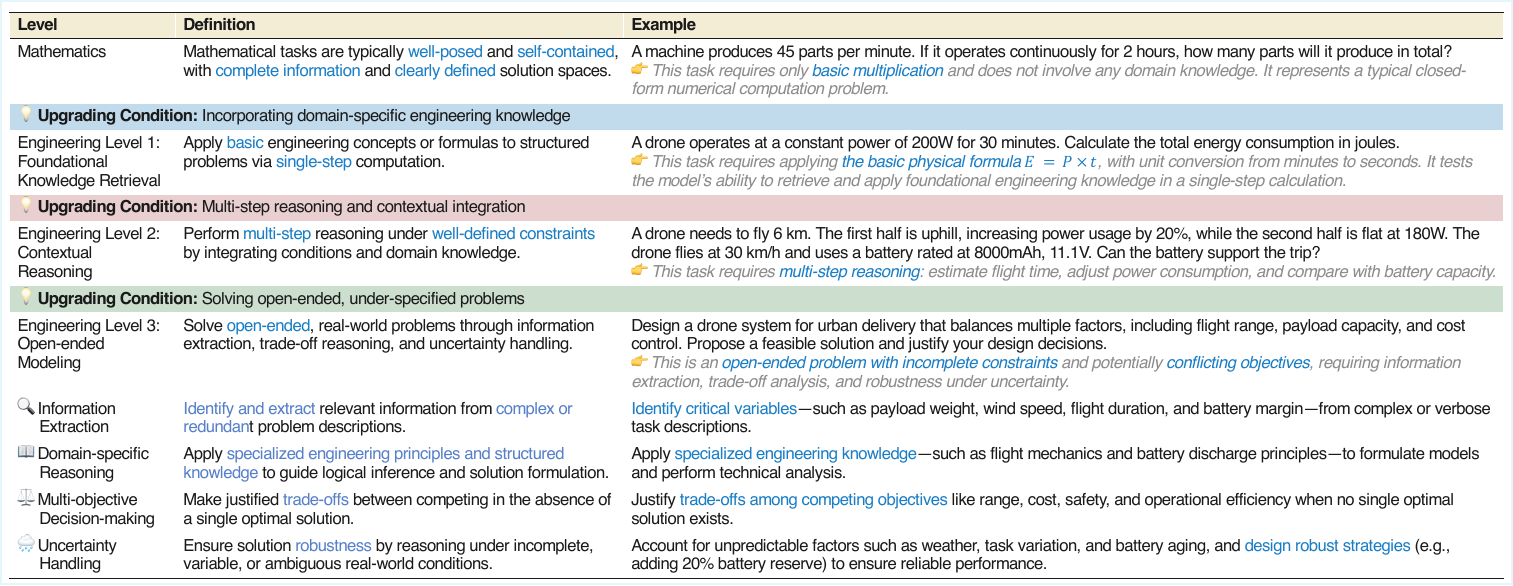}  
\caption{Hierarchical difficulty from mathematics to real-world engineering. This illustrates three levels of increasing complexity. Examples show the progression from closed-form math problems to open-ended engineering scenarios.}
\label{tab:Task hierarchy}
\vspace{-2mm}
\end{table*}

\section{Related Works}

\noindent \textbf{LLMs for Engineering Problems.}\quad
LLMs integrate broad domain knowledge with strong multi-step reasoning, making them promising tools for complex problem solving.
Engineering problems, however, require modeling real-world systems and reasoning under practical constraints.
Despite the growing use of LLMs in engineering applications, their true engineering problem-solving capability remains unclear due to the limitations of existing benchmarks \citep{wang2024news, ma2024llm,tang2024orlm,cheng2025large}.
General-purpose benchmarks, including MMLU \citep{hendrycks2021measuring}, MMLU-Pro \citep{wang2024mmlu}, BIG-Math \citep{albalak2025big}, and SuperGPQA \citep{du2025supergpqa}, contain only limited engineering content.
Most questions emphasize factual recall in multiple-choice form, failing to capture core engineering reasoning.
Several domain-specific engineering benchmarks have been proposed, including EEE-Bench \citep{li2024eee}, ElecBench \citep{zhou2024elecbench}, FEABench \citep{mudur2025feabench}, TransportBench \citep{syed2024benchmarking}, and JEEBench \citep{arora-etal-2023-llms}.
However, they largely focus on single disciplines and well-defined tasks, limiting their ability to evaluate open-ended and cross-disciplinary engineering reasoning.
To address this gap, we introduce a multi-level engineering benchmark spanning multiple subfields and incorporating both closed-form and open-ended tasks, enabling a comprehensive evaluation of engineering capabilities.

\noindent \textbf{LLM for Mathematical Problems.}\quad 
A closely related area that has been extensively studied is mathematics.
Because solving mathematical problems demands strong logical ability, multi-step reasoning, and symbolic manipulation, it has become a primary proving ground for evaluating LLMs.
Early benchmarks focus on elementary problems \citep{cobbe2021training, hendrycks2021measuring,patel2021nlp,amini2019mathqa} and  higher-level symbolic reasoning \citep{hendrycks2021measuring,albalak2025big}.
Recent efforts like MiniF2F \citep{zheng2022miniff}, UniMath \citep{liang2023unimath}, Omni-MATH \citep{gao2024omni}, and MathVista \citep{lu2024mathvista} expand to theorem proving and multimodal tasks. MATH-Vision \citep{wang2024measuring} improves coverage by introducing diverse topics and difficulty levels from real competitions, and SMART-840 \citep{cherian2024evaluating} benchmarks model performance against human children across grades.
While these benchmarks provide rigorous evaluations of mathematical competence, they do not capture engineering-specific reasoning such as modeling, decision-making under constraints, or domain-based assumptions.
Our work builds on their methodological insights but shifts the focus toward real-world engineering tasks.

\noindent \textbf{Evaluation Challenges.}\quad
Evaluating the capability of LLMs to solve engineering problems is challenging due to the inherent complexity involved.
Current evaluation methods for LLMs fall into four main categories: reference-based, task-oriented, preference-based, and rubric-based.
The first two are effective for problems with clear ground truths or executable outputs -- e.g., MathVista \cite{lu2024mathvista}, CHAMP \cite{mao-etal-2024-champ} (reference-based), and EEE-Bench \cite{li2024eee}, FEABench (task-oriented) \cite{mudur2025feabench}.
However, the core capabilities of the engineering field we are discussing cannot be effectively evaluated by such closed-form problems.
For open-ended tasks, preference-based methods such as MT-Bench-101 \cite{bai2024mt} use pairwise comparisons, but are often biased by model-specific generation patterns, limiting objectivity and real-world applicability.
Rubric-based evaluations aim to improve transparency by scoring along multiple criteria, with general-purpose frameworks like Prometheus \cite{kim2024prometheus} focusing on abilities such as context retention and rephrasing.

\section{Methodology}
\label{sec: EngiBench}

\subsection{Engineering Problem-Solving Capability}\label{sec: Engineering Problem-Solving Capability}

Engineering problems require context-aware solutions under real-world constraints, distinguishing them from mathematical problems that typically operate in well-defined, closed-form settings \citep{dym2005engineering,hendrycks2021measuring}. In this work, engineering problems are defined as tasks that apply scientific principles to the modeling and analysis of systems under such constraints.
Beyond abstraction and logical rigor, engineering problem solving involves a sequence of interconnected cognitive steps, from interpreting problem context to making decisions under constraints and uncertainty.
We refer to this as engineering problem-solving ability, which comprises four key dimensions: information extraction, domain-specific reasoning, multi-objective decision-making, and uncertainty handling.
These dimensions align with established paradigms in engineering modeling, including information filtering, constraint-based and multi-objective formulation, and robustness analysis, and can be interpreted as a reasoning-level abstraction of classical engineering design processes \citep{beitz1996engineering}, spanning stages such as task clarification, conceptual design, embodiment design, and detail design (see Figure~\ref{fig_0} and Table~\ref{tab:Task hierarchy}).

\noindent \textbf{Information Extraction.}\quad The capability to identify and organize key variables, constraints, and objectives from complex or noisy descriptions. It reflects the model’s capacity to filter irrelevant information and transform unstructured text into structured representations for downstream reasoning.

\noindent \textbf{Domain-specific Reasoning.}\quad The capability to apply engineering knowledge, including physical principles, empirical rules, and domain conventions, to interpret scenarios and choose appropriate solution strategies. It involves recognizing valid approximations, implicit assumptions, and methods used in engineering practice.
    
\noindent \textbf{Multi-objective Decision-making.}\quad The capability to balance competing objectives such as cost, performance, and safety when no single optimal solution exists. This dimension assesses whether a model can justify trade-offs under constraints, a defining feature of engineering problem solving.
    
\noindent \textbf{Uncertainty Handling.}\quad The capability to reason under incomplete, noisy, or dynamic information. It includes anticipating uncertainties, incorporating safety margins or fallback strategies, and generating solutions that remain robust despite ambiguity. This capability is essential for making reliable engineering decisions in real-world settings.

\subsection{Problem Hierarchical Difficulty Design}\label{sec: Difficulty Levels}
Engineering problem solving involves multiple distinct capabilities, making it difficult to assess through a single task or a one-dimensional hierarchy. A clear taxonomy is therefore essential for identifying where models succeed or fail. To provide such structure, EngiBench organizes engineering tasks into three complementary levels: foundational knowledge retrieval, contextual reasoning, and open-ended modeling, each reflecting different cognitive demands. Rather than simply aggregating tasks, this framework organizes evaluation by reasoning complexity, forming a hierarchy consistent with Bloom’s Taxonomy \citep{krathwohl2002revision}. 

\noindent \textbf{Level 1.}\quad
Tasks are well-defined and self-contained, typically requiring only single-step application of fundamental engineering formulas.
They emphasize factual recall, accurate computation, and minimal contextual reasoning.
This level assesses whether a model has a stable engineering knowledge base and can reliably retrieve and apply it to straightforward problems.

\noindent \textbf{Level 2.}\quad
Tasks require multi-step reasoning under contextual constraints such as units, physical limits, and coupled variables.
Although these problems are well-defined and have unique solutions, models must interpret structured descriptions and integrate domain knowledge across steps to generate correct answers.
Compared with Level 1, simple recall is insufficient; models need to handle structured complexity to generate correct solutions.

\noindent \textbf{Level 3.}\quad  
Tasks reflect open-ended engineering challenges with uncertainty, incomplete information, and conflicting objectives. They require the full engineering problem-solving capability. Unlike Level 1 and Level 2, problems do not have a single correct answer, and evaluation focuses on how well models demonstrate robust and adaptive reasoning under open-ended conditions.

\subsection{Dataset Construction}
\label{sec: Data Generation and Structure}
\noindent \textbf{Data Sources.}\quad
We collect data from three primary sources: problems selected from existing public benchmarks, university educational materials, and modeling competitions.
These problems reflect the intended hierarchy of difficulty described above and address the lack of open-ended engineering modeling problems with expert-defined evaluation criteria in existing datasets.

\noindent \textbf{Construction Process.}\quad 
Levels 1 and 2 contain structured engineering problems with standard answers, drawn from general-domain benchmarks such as SuperGPQA \citep{du2025supergpqa}, MMLU \citep{hendrycks2021measuring}, MATH \citep{hendrycks2021measuring}, GSM8k \citep{cobbe2021training}, Orca-Math \citep{mitra2024orca}, HARP \citep{yue2024harp}, Omni-MATH \citep{gao2025omnimath}, Big-Math \citep{albalak2025big}, and selected university resources. Although these datasets are broad in scope, all problems used in EngiBench were passed through an engineering relevance filtering procedure (Appendix \ref{appsec: Extraction Process}) to retain only questions that align with engineering knowledge. All selected problems were further standardized and validated.

Level 3 introduces the first systematic collection of open-ended engineering tasks, comprising 43 problems from major modeling competitions. Each problem includes official scoring rubrics and reference solutions provided by top-ranking competition winners.
All task rewrites and scoring rubrics were finalized by domain experts, with LLMs providing auxiliary support during intermediate steps, ensuring clarity, rigor, and reliable assessment (see Appendix \ref{sec: Level 3 Data Collection and Processing} and Appendix \ref{sec: Level 3 Evaluation Details}).

\noindent \textbf{Problem Annotation and Quality Control.}\quad Level 3 problems and their scoring rubrics were expert-reviewed by 20 PhD students and engineering professionals, with LLMs used only as auxiliary tools. From nearly 1,000 competition questions, we retained only those with official rubrics and performed extensive text-based reformulation of formulas, tables, and diagrams. The released scoring scripts implement these expert-defined rubrics for reproducible downstream evaluation. All annotation and quality control in this section are limited to problem and rubric construction (see Appendix~\ref{sec: Rubric Construction}), while model response scoring is described in Section~\ref{sec: Experiment Setup}.

\noindent \textbf{Coverage and Classification.}\quad EngiBench spans three subfields: Systems \& Control (939 problems), Physical \& Structural (354 problems), and Chemical \& Biological (467 problems). This categorization reflects differences in problem focus, underlying domain knowledge, and the reasoning processes required to solve them.

\subsection{Controlled Problem Variants}

Evaluating LLMs on engineering tasks requires more than measuring overall accuracy. A correct answer may arise from data memorization rather than reasoning  \citep{huang2025math, zhang2024careful, mirzadeh2025gsmsymbolic, srivastava2024functional, gulati2024putnam}, while an incorrect answer may reflect missing domain knowledge, weak mathematical skills, or failures in interpreting engineering constraints. Without separating these factors, accuracy alone provides limited diagnostic value.

To enable deeper analysis, each problem is rewritten into three controlled variants derived from the original form (Figure \ref{fig:problem_version}). (1) The \textit{perturbed variant} introduces numerical and semantic changes to assess robustness and reduce possible overlap with pretraining data. (2) The \textit{knowledge-enhanced variant} adds essential domain information such as formulas, constants, and key definitions so that errors caused by missing knowledge can be distinguished from reasoning failures. (3) The \textit{math abstraction variant} removes contextual and domain-specific elements while preserving the underlying mathematical structure, allowing us to isolate mathematical reasoning and quantify how much engineering context affects performance.

\begin{figure}[t]
\label{fig:problem_version}
\centering
\includegraphics[width=0.5\textwidth]{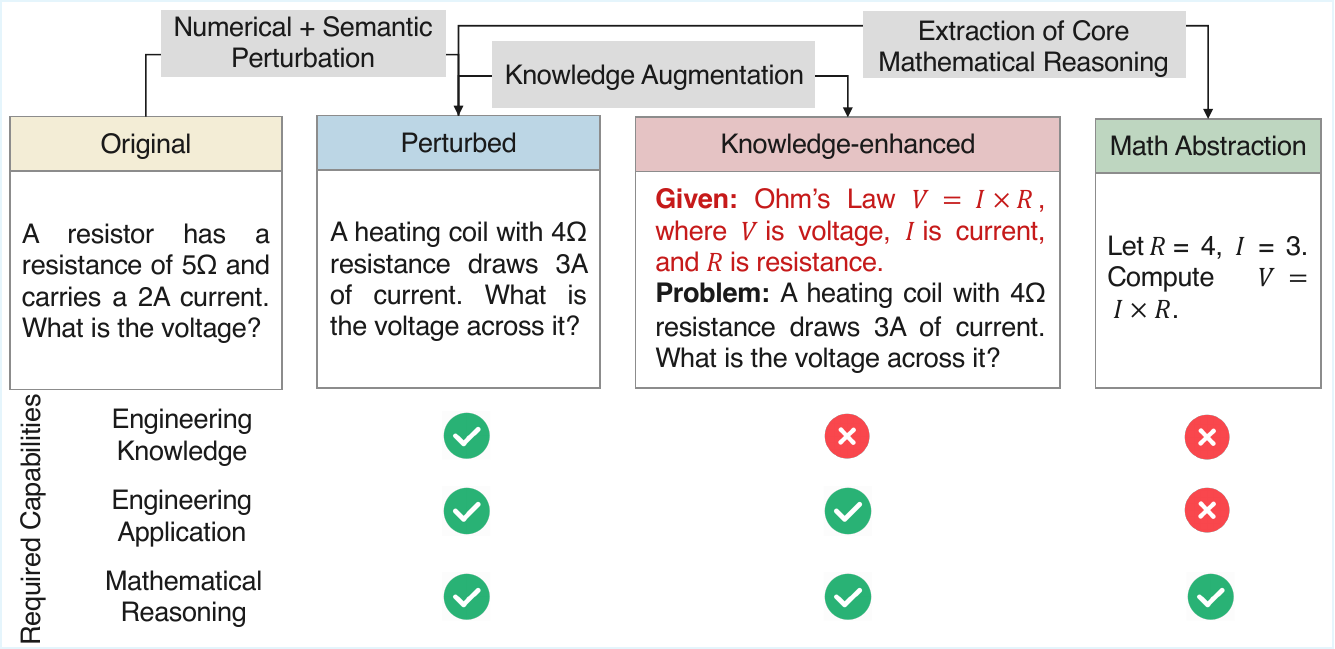}  \label{fig:problem_version}
\caption{We create variants of the original problem to test different reasoning skills. \textit{Perturbed} changes context and numbers to assess robustness. \textit{Knowledge-enhanced} adds domain knowledge to focus on reasoning. \textit{Math Abstraction} isolates engineering knowledge to test math ability. Each version targets specific capabilities.
}
\label{fig:problem_version}
\end{figure}

These controlled variants provide a structured way to distinguish why a model succeeds or fails, giving a capability-oriented evaluation of engineering problem solving. For Level 1 and Level 2, all three variants are constructed systematically from the original problem. For Level 3, the open-ended nature and inherent complexity make knowledge-enhanced and math abstraction variants impractical, so only the perturbed variant is included.

\begin{figure*}[t]
\centering
\captionsetup{font={small}, skip=8pt}
\includegraphics[width=\textwidth]{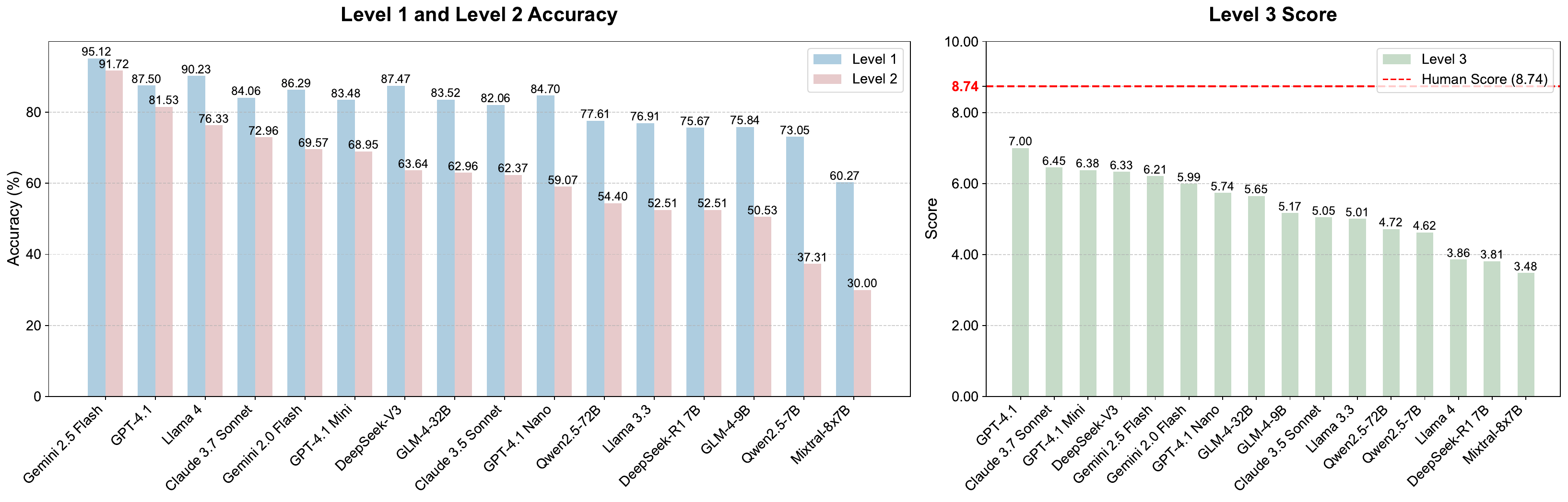}
\caption{Overview of model performance across engineering reasoning tasks. The left subfigure shows model accuracy on Level 1 and Level 2 tasks, while the right subfigure presents scores on Level 3 open-ended tasks, with the human expert score indicated by the red line.}
\label{fig: total}
\end{figure*}

\section{Experiments}

\subsection{Experiment Setup}\label{sec: Experiment Setup}

\textbf{Evaluated LLMs.}\quad As the first batch, 16 LLMs were evaluated under the zero-shot setting, covering a representative range of model types.
Specifically, we include:
(1) closed-source models such as GPT-4.1, GPT-4.1 Mini, and GPT-4.1 Nano from OpenAI \citep{achiam2023gpt}; Claude 3.7 Sonnet and Claude 3.5 Sonnet from Anthropic \citep{anthropic_claude3_5_sonnet_2024, anthropic_claude3_family_2024}; and Gemini 2.5 Flash and Gemini 2.0 Flash from Google DeepMind \citep{team2023gemini, team2024gemini};
(2) open-source models, including GLM-4-32B and GLM-4-9B from THUDM \citep{glm2024chatglm}, Qwen2.5-72B and Qwen2.5-7B from Alibaba \citep{yang2024qwen2}, Llama 4 Maverick (referred to as Llama 4) and Llama 3.3-70B (referred to as Llama 3.3) from Meta \citep{grattafiori2024llama}, and DeepSeek-V3-671B (referred to as DeepSeek-V3) and DeepSeek-R1-Distill-Qwen-7B (referred to as DeepSeek-R1 7B) from DeepSeek \citep{guo2025deepseek}, Mixtral-8x7B-Instruct-v0.1 (referred to as Mixtral 8x7B) from Mistral AI \citep{jiang2024mixtral}.
This selection spans diverse model sizes, training paradigms, and accessibility levels. We ensured consistent formatting and output parsing across all models.

\noindent \textbf{Evaluation protocols.}\quad 
Level~1 and Level~2 consist of well-defined problems with unique solutions and are evaluated using binary scoring. Evaluation consistency is verified through multi-model cross-checking and random human spot checks. Further details are provided in Appendix~\ref{sec: Level 1 Level 2 Evaluation Details}.
Level~3 tasks are open-ended and are evaluated using a rubric-based framework derived from official criteria and refined by domain experts. Scoring is performed by LLMs following the same rubrics, and all Level~3 scores are subsequently reviewed and calibrated by human annotators following the same criteria. Further details are provided in Appendices~\ref{sec: Rubric-based Scoring and Human Calibration} and \ref{sec: Level 3 Scoring Consistency Analysis}.

Also, we introduce human scores for Level 3 tasks for comparison with LLMs' performance.
We obtain human scores from two sources: award-winning competition submissions (original version) and manual solutions by top-performing students for the perturbed variant.
All human and LLM responses are evaluated using the same rubric to ensure consistency and fairness.

\begin{figure*}[t]
\centering
\captionsetup{font={small}, skip=8pt}
\includegraphics[width=1.0\textwidth]
{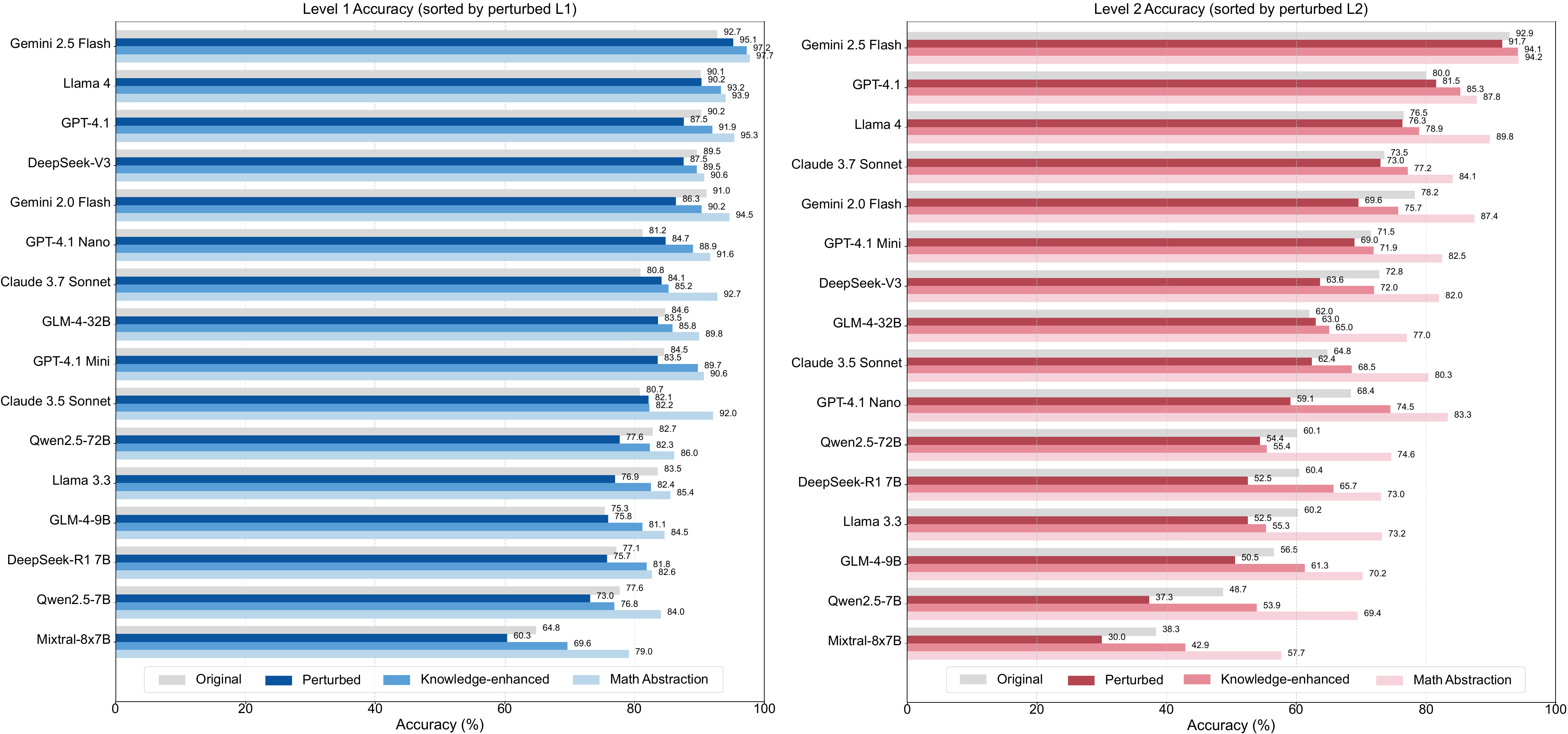}
\caption{Accuracy of LLMs on Level~1 (left) and Level~2 (right) across the Original, Perturbed, Knowledge-enhanced, and Math Abstraction variants. Drops under the Perturbed variant reflect sensitivity to input changes, while gains on the latter two indicate that models benefit from added knowledge or simplified formulations.
}
\label{fig: level1&2}
\end{figure*}

\subsection{Results}
\subsubsection{Overall}\label{sec:overall}

\textbf{Model stratification and design validation.}\quad 
Model performance exhibits a clear downward trend from Level 1 to Level 3, demonstrating the effectiveness of our hierarchical difficulty design. As shown in Figure~\ref{fig: total}, most models achieve high accuracy on Level 1, perform moderately on Level 2, and show a clear performance decline on Level 3. This trend indicates that our hierarchical framework successfully separates problems by cognitive difficulty, with each level reflecting distinct capability thresholds. The results validate that a multi-level design is necessary to capture the full range of engineering problem-solving capabilities.

\noindent \textbf{Evaluating high-level engineering reasoning.}\quad 
Level 3 is designed to assess high-level engineering reasoning that goes beyond formulaic computation. Unlike Level 1 and Level 2, which focus on structured problem solving, Level 3 features open-ended and underspecified tasks that better reflect real-world engineering challenges. The sharp performance drop at this level reveals the current limitations of LLMs in handling such complex scenarios. 
Besides, the gap between LLMs and human experts at Level 3 also reveals a key deficiency in high-level engineering capabilities. All evaluated models score well below the human expert, who achieves an average of 8.74, indicating that current LLMs are still far from reliably handling complex engineering problems. This underscores the need for further research to bridge this gap. 

\noindent \textbf{Smaller-scale LLMs struggle with complex tasks.}\quad 
While all LLMs show room for improvement on complex, open-ended engineering tasks, smaller-scale LLMs exhibit significantly greater limitations.
As task complexity increases, performance disparities widen.
At Level 1, most models still cluster within 70--90\%.
But at Level 2, leading models such as GPT-4.1 and Gemini~2.5 Flash achieve accuracies above 80\%, whereas DeepSeek-R1~7B reaches only about 52\% and other lightweight models often fall below 40\%.
This divergence is most evident at Level~3, where state-of-the-art models approach scores of 7.0, while lightweight models remain under 4.0.
These results show that EngiBench is not saturated and continues to distinguish models across scales.

\noindent \textbf{Robustness and contamination risk.}\quad 
Some LLMs may achieve high scores not through internal reasoning, but due to overlap with pretraining data. To reveal this, we use a perturbed variant that applies minor contextual and numerical changes but keep the core structure unchanged. As shown in Figure~\ref{fig: level1&2}, model performance remains relatively stable on Level 1 but drops sharply on Level 2. For example, on Level 2, accuracy decreases by 9.3\% for GPT-4.1 Nano, 11.4\% for Qwen2.5-7B, and 8.3\% for Mixtral-8x7B. These declines suggest a stronger reliance on surface-level pattern matching, rather than robust reasoning, highlighting the role of perturbation-based evaluation in diagnosing overestimated capabilities.

\begin{figure*}[t]
  \centering
  \begin{subfigure}[t]{0.56\textwidth}    \includegraphics[width=\linewidth]{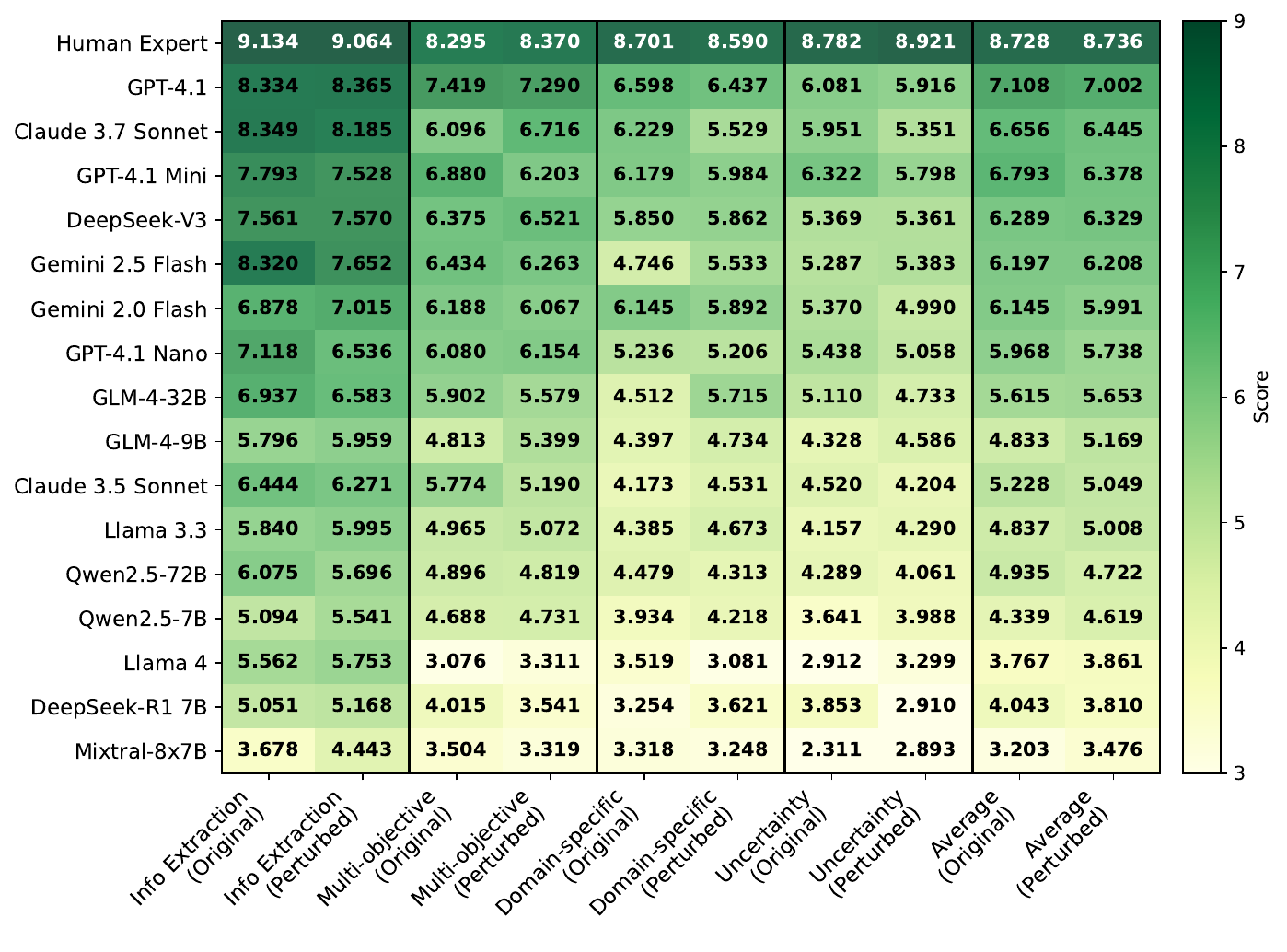}
    \caption{Level 3 Model Evaluation. }
    \label{fig:level3}
  \end{subfigure}
  \hfill
  \begin{subfigure}[t]{0.42\textwidth}
\includegraphics[width=\linewidth]{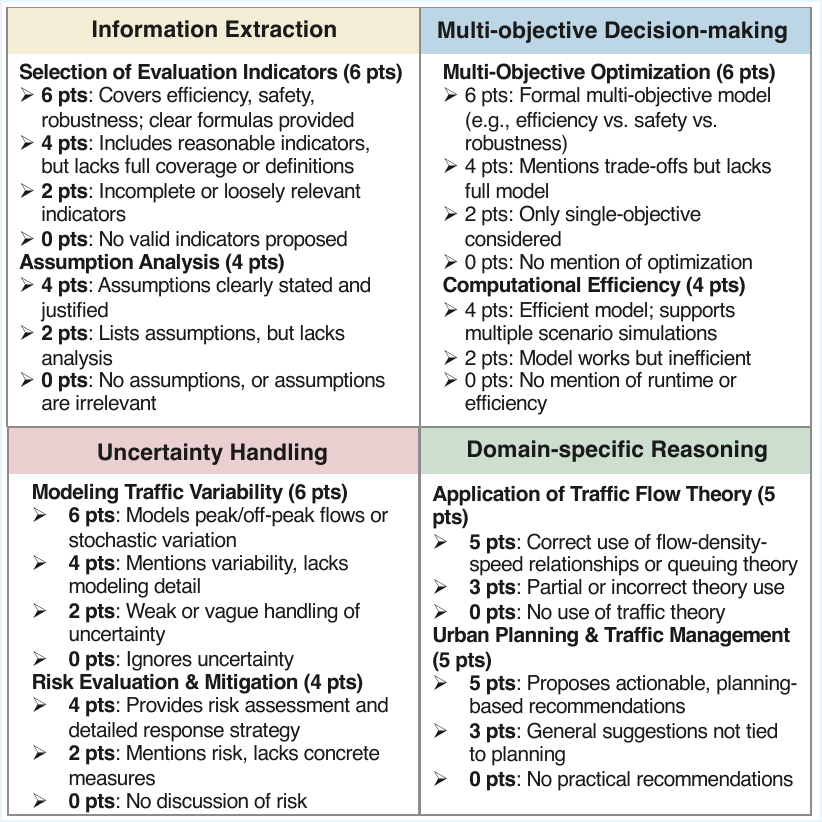}
    \caption{Scoring rubric example.}
    \label{fig:level3_metrics}
  \end{subfigure}
  \caption{Level 3 Model Evaluation and Scoring Rubric. This figure summarizes Level 3 evaluation results and scoring standards. Subfigure (a) reports average model scores across four capabilities under both original and perturbed inputs. Subfigure (b) shows an example rubric outlining scoring criteria across capability dimensions.
}
\label{fig:combined_figures}
\end{figure*}

\begin{figure*}[t]
  \centering
  \begin{minipage}[t]{0.4\textwidth}
    \centering
    \includegraphics[width=\linewidth]{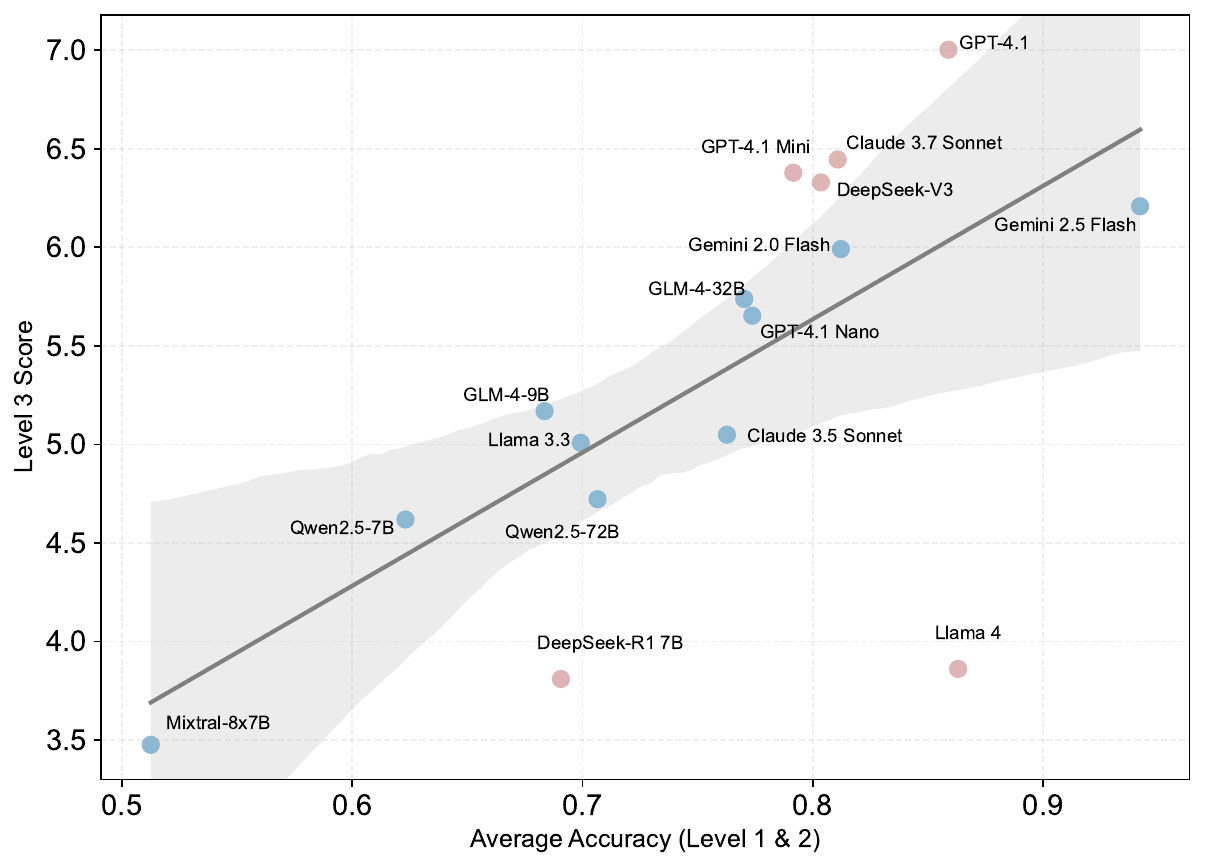}
    \caption{Correlation between structured tasks (Level 1\&2) and open-ended tasks (Level 3). 
    }
    \label{fig:correlation}
  \end{minipage}%
  \hfill
  \begin{minipage}[t]{0.55\textwidth}
    \centering
    \includegraphics[width=\linewidth]{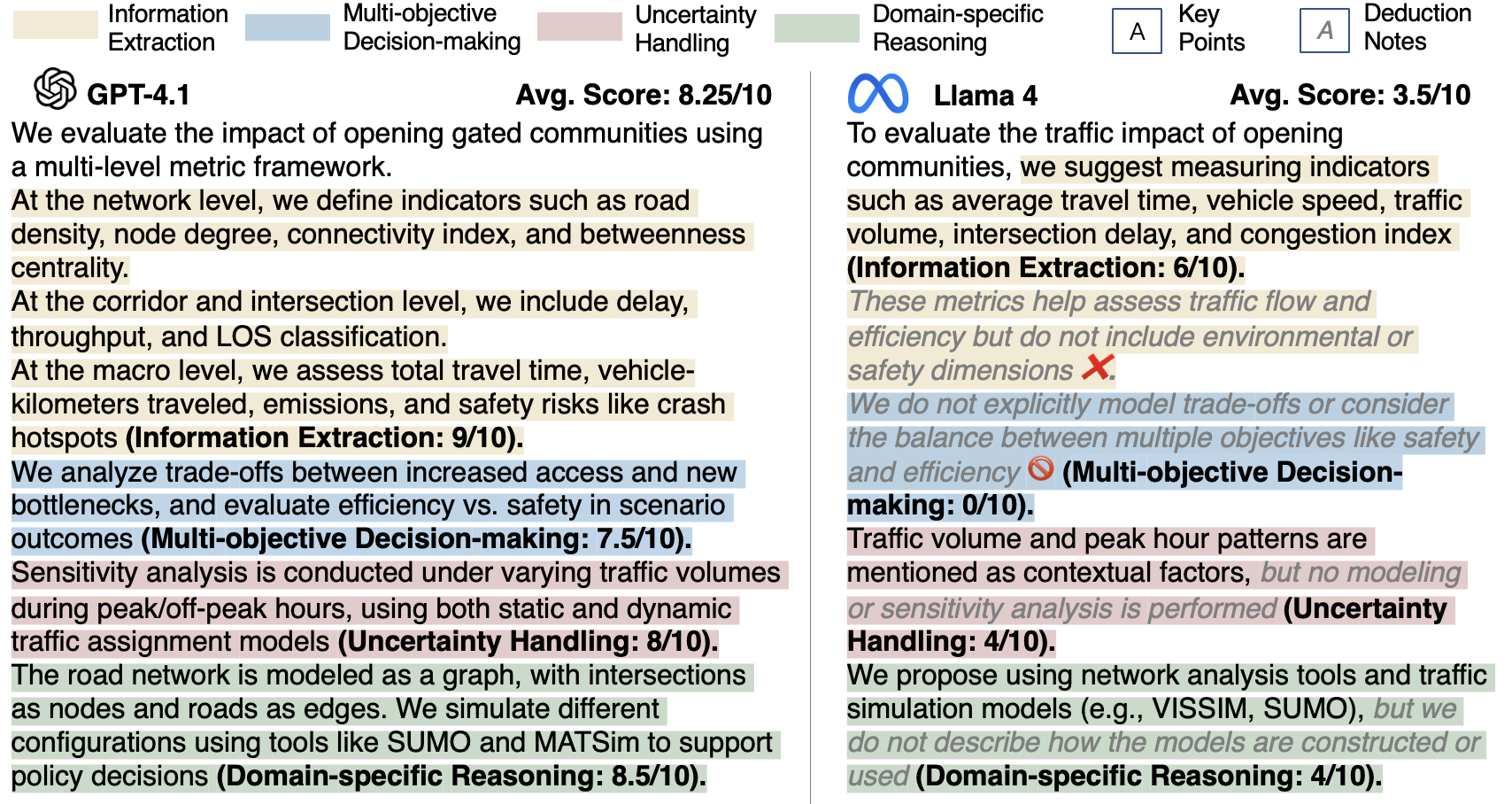}
    \caption{Case study showing why Llama 4 received low Level 3 scores. 
    }
    \label{fig:llama4}
  \end{minipage}
\end{figure*}

\subsubsection{Performance for Level 1 \& Level 2 Tasks}

\noindent \textbf{Knowledge Enhancement Improves Accuracy.}\quad
Adding explicit domain knowledge consistently improves accuracy across all levels, especially for weaker models. As shown in Figure~\ref{fig: level1&2}, models perform better on knowledge-enhanced variants than on perturbed inputs. These gains suggest two main failure sources: lacking essential domain knowledge or failing to apply it correctly during reasoning. Providing explicit knowledge therefore offers a clear diagnostic signal that helps distinguish knowledge deficits from reasoning errors, which is a key capability for engineering evaluation.

\noindent \textbf{Math Abstraction Further Improves Performance.}\quad
LLMs perform even better when engineering problems are rewritten into purely mathematical form, removing contextual details. As shown in Figure~\ref{fig: level1&2}, most models achieve their highest accuracy under this variant, especially smaller models that struggle with contextual interpretation. This pattern suggests that the main challenge in engineering tasks is not computation, but the earlier step of translating natural-language descriptions into well-structured mathematical formulations.
This underscores the importance of evaluating the reasoning steps that precede formula application, as these upstream processes are not captured by traditional math benchmarks.

\noindent \textbf{Smaller Models Are More Sensitive to Input Variants.}\quad
Smaller-scale LLMs exhibit much larger performance fluctuations across input versions, indicating limited generalization and unstable reasoning. As shown in Figure~\ref{fig: level1&2}, in Level 2, Qwen2.5-7B drops by 11.4\% under the perturbed variant, yet gains 16.6\% with added domain knowledge and another 15.5\% under math abstraction. In contrast, Gemini~2.5 Flash remains highly stable: its accuracy decreases by only 1.2\% under the perturbed version and increases by 2.4\% and 2.5\% under the knowledge-enhanced and math abstraction variants, respectively. This comparison shows that smaller models are more sensitive to input formulation and tend to rely on surface patterns rather than consistent, context-aware reasoning.

\subsubsection{Performance for Level 3 Tasks}\label{sec: Performance for Level 3 Tasks}

\textbf{Dimension-wise and model-wise performance.}\quad As shown in Figure~\ref{fig:level3}, human experts lead across all four dimensions with a balanced capability profile. In contrast, LLMs show uneven performance across the four dimensions. They handle redundant information extraction relatively well and perform moderately on multi-objective decision-making but struggle with domain-specific reasoning and uncertainty handling. This pattern indicates that their abilities are imbalanced, with clear deficiencies in key engineering-oriented skills. Results also demonstrate that model performance correlates with scale and accessibility. Larger, closed-source models like GPT-4.1 and Claude 3.7 Sonnet, consistently achieve average scores above 6. In contrast, smaller open-source models (e.g., Mixtral-8x7B) average below 4, with common omissions in aspects such as trade-off reasoning and uncertainty consideration.

\noindent \textbf{Correlation analysis.}\quad To quantify this trend, Figure~\ref{fig:correlation} illustrates the relationship between model performance on structured tasks (Levels 1 \& 2) and open-ended tasks (Level 3). Overall, we observe a clear positive correlation: \textit{models that achieve higher accuracy on structured tasks tend to also perform well on open-ended tasks}, suggesting a general consistency across task types.
At the same time, some models deviate from this general trend. GPT-4.1, Claude 3.7 Sonnet, and DeepSeek-V3 show notably stronger performance on Level 3 than their results on Levels 1 and 2 would suggest, indicating more advanced reasoning and modeling abilities than what structured tasks alone reveal.

In contrast, models like Llama 4 perform pretty well on structured tasks but falter on open-ended ones, revealing weak high-level reasoning. Figure~\ref{fig:llama4} illustrates this gap: Llama 4 scores 0 in multi-objective decision-making due to missing trade-off analysis, while GPT-4.1 provides a structured evaluation and scores 7.5. A similar shortfall also appears in uncertainty handling. These examples show that Llama 4 can recall facts but struggles to apply them in complex, judgment-based scenarios.

\section{Conclusion}

We introduce \textbf{EngiBench}, a benchmark for evaluating LLMs on engineering problem solving across increasing levels of complexity. Our results show that while current models perform well on foundational knowledge retrieval, their performance declines significantly in multi-step contextual reasoning tasks, due to both domain knowledge gaps and limited mathematical reasoning. On open-ended modeling tasks, even the strongest models fall short of human-level performance, revealing persistent limitations in high-level reasoning, trade-off analysis, and uncertainty handling. These findings underscore the need for LLMs to move beyond pattern matching and toward deeper reasoning capabilities for real-world engineering applications.

\section*{Limitations}
While EngiBench provides the first systematic evaluation of LLMs on real-world engineering problems, covering multi-level tasks, variant-based reasoning diagnostics, and open-ended modeling, several limitations remain that we plan to address in future work.

\noindent\textbf{Multimodal Support.}\quad
Many real-world engineering problems involve visual elements such as diagrams, schematics, or structured tables. The current version of EngiBench does not include multimodal tasks, as most existing LLMs still lack stable and consistent multimodal input capabilities. To avoid confounding engineering reasoning performance with visual processing variability and to ensure fair and comparable evaluation across models, we restrict all inputs to text-only formats.

\noindent\textbf{Long-Context Support.}\quad
Some engineering tasks involve long problem descriptions or extensive tabular data that exceed the input length limits of current LLMs. To avoid unfair model truncation effects and ensure uniform evaluation settings, such problems are excluded from this version.

\noindent\textbf{Human-in-the-loop Construction.}\quad
Building the dataset involves substantial human effort, including problem collection, answer generation, and variant validation. This ensures data quality and alignment with engineering standards, but also reflects the significant manual effort behind the benchmark.

\section*{Acknowledgements}
This work was supported in part by the Ministry of Education and Science of Bulgaria (support for INSAIT, part of the Bulgarian National Roadmap for Research Infrastructure), the Shenzhen Institute of Artificial Intelligence and Robotics for Society (AIRS), the Shenzhen Key Laboratory of Crowd Intelligence Empowered Low-Carbon Energy Network (No.~ZDSYS20220606100601002), the National Natural Science Foundation of China (No.~72331009), and the Ministry of Education (MOE), Republic of Singapore, under Grant AcRF Tier~1 (RG59/22).

We would like to express our sincere gratitude to Mo Chen, Zixuan Cui, Rui Jin, Kaicheng Li, Zhuoqi Li, Jili Tu, Bihua Wen, and Jiaxiang Xie for their valuable contributions to the construction, annotation, and validation of the EngiBench.

\bibliography{custom}
\clearpage

\newpage
\appendix

\let\addcontentsline\oldaddcontentsline

\section*{Appendix}
\tableofcontents

\section{The Use of Large Language Models}

In this work, LLMs were used in three ways: (1) grammar checking and language polishing during paper writing, (2) generating controlled problem variants in the benchmark construction process, and (3) serving as both the models under evaluation and auxiliary judges for rubric-based scoring.

\section{Ethical Considerations}
This work introduces a benchmark for evaluating large language models on engineering tasks. The problems are derived from publicly available benchmarks, academic competitions, and educational materials. For open-ended tasks, human participants voluntarily contributed reference solutions and evaluation scores using publicly available rubric criteria, and personal information was collected only for inclusion in the acknowledgment section with explicit consent. The dataset does not contain sensitive data or enable harmful applications. 
EngiBench is designed as an evaluation framework for systematically analyzing and comparing model behaviors across diverse engineering task settings.
The goal of EngiBench is to promote rigorous, transparent, and fair evaluation of language models in engineering contexts, and we affirm adherence to the ACL Code of Ethics, including principles of fairness, transparency, and research integrity.

\section{Future Works}
While EngiBench establishes a strong foundation for evaluating LLMs on engineering problem-solving, several avenues remain for further development and expansion:

\noindent \textbf{Scalability Across Engineering Domains.}\quad
EngiBench currently covers three core engineering subfields—Systems \& Control, Physical \& Structural, and Chemical \& Biological—which together span a wide range of disciplines such as Mechanical, Electrical, and Chemical/Biological Engineering. The benchmark framework is designed to be broadly applicable and adaptable across domains. In future work, we plan to expand the dataset by incorporating problems from additional engineering disciplines to further enhance data volume and subject diversity.

\noindent \textbf{Multimodal Evaluation Extensions.}\quad
Future versions of EngiBench will introduce a dedicated multimodal subset to evaluate models on tasks involving vision-language reasoning. This will enable systematic assessment of model performance in scenarios that demand visual interpretation alongside textual understanding.

\noindent \textbf{Support for Long-Context Reasoning.}\quad
We plan to extend the benchmark to include long-context engineering tasks by leveraging models with expanded context windows or hierarchical processing capabilities. This will allow for evaluation of more complex, information-rich tasks currently excluded due to input length limitations.

\section{Dataset Construction}

\subsection{Level 1 \& Level 2 Extraction Process}
\label{appsec: Extraction Process}

To construct a high-quality and diverse dataset for Level~1 and Level~2, we systematically extract relevant tasks from a range of established public benchmarks, including MMLU \citep{hendrycks2021measuring}, MATH \citep{hendrycks2021measuring}, GSM8k \citep{cobbe2021training}, Orca-Math \citep{mitra2024orca}, HARP \citep{yue2024harp}, Omni-MATH \citep{gao2025omnimath}, Big-MATH \citep{albalak2025big}, and competition datasets such as cn\_k12, Olympiads, AOPS forum, and AMC-AIME \citep{li2024bCompetitionMath}. In addition to these public sources, we also incorporate university-level engineering educational materials, including assignments, examinations, and instructor-provided teaching content, to further increase task diversity and real-world relevance.

To transform mathematical and logic-oriented problems into engineering-relevant evaluation tasks, we design a structured data processing pipeline that combines LLM-based analysis with human verification to ensure engineering relevance and classification accuracy. This pipeline ensures that all included problems align with real-world engineering semantics and reasoning demands, forming the basis for Level~1 and Level~2 in EngiBench. 

The processing pipeline consists of the following steps:

\begin{enumerate}
    \item \textbf{Engineering Relevance Filtering:} Each problem is evaluated for its applicability to engineering scenarios. Problems lacking domain relevance are excluded to maintain the technical integrity of the benchmark. The prompt used to determine whether a problem pertains to engineering is as follows:
    \begin{lstlisting}
    """Determine if ORIGINAL problem can be solved with ONLY mathematical knowledge (NO engineering background):
   - False if requires any domain-specific knowledge
   - True if solvable through pure mathematical calculations"""
    \end{lstlisting}
    
    \item \textbf{Discipline and Subfield Classification:} Relevant problems are first assigned to a specific engineering discipline (e.g., Electrical, Civil, Mechanical), and then grouped into one of EngiBench’s three high-level analytical subfields: Systems \& Control, Physical \& Structural, or Chemical \& Biological. The prompt used for assigning a problem to a specific engineering discipline is as follows:
    \begin{lstlisting}
    """If yes, which engineering category? (Chemical/Bioengineering/Geotechnical/Energy/Nuclear/Aerospace/Automotive/Biomedical/Civil/Control/Electrical/Industrial/Mechanical/Ocean/Environmental/Other) (Please try to avoid Other)
    If not an engineering problem, return "N/A"."""
    \end{lstlisting}
    
    \item \textbf{Difficulty Level Assignment:} Based on the complexity of the required reasoning process, tasks are categorized into Level 1 or Level 2. Level 1 includes basic knowledge recall and single-step computation, while Level 2 involves multi-step inference, contextual understanding, and integration of structured constraints. The prompt used for classifying the difficulty level of a problem is as follows:
    \begin{lstlisting}
    """Difficulty level? (Level 1/Level 2) (Please try to avoid unknown):
   - Level 1: The problem can be solved by a direct retrieval of information or by directly substituting values into a known formula—i.e., the shortest possible solution path. No chaining of intermediate steps is required. (Example: Using Ohm's Law, V = IR, to directly compute voltage when given current and resistance.)
   - Level 2: The problem requires multi-step reasoning—meaning that it involves chaining together several logical deductions, intermediate calculations, or systematic strategies beyond a single direct formula application. (Example: Analyzing a circuit to compute total resistance by first calculating individual branch resistances and then combining them.)"""
    \end{lstlisting}
\end{enumerate}

\subsection{Level 3 Data Collection and Processing}\label{sec: Level 3 Data Collection and Processing}

To construct the Level~3 dataset in \textbf{EngiBench}, we focus on real-world, open-ended engineering tasks sourced from major mathematical modeling competitions. Specifically, we collect problems from publicly accessible archives of contests such as the China Undergraduate Mathematical Contest in Modeling (CUMCM), the Mathematical Contest in Modeling / Interdisciplinary Contest in Modeling (MCM/ICM), and the Asia and Pacific Mathematical Contest in Modeling (APMCM), covering the years 2010 to 2024.

To ensure domain relevance and evaluation consistency, we apply strict filtering criteria. We retain only problems with clear engineering context and official scoring rubrics, and exclude those that depend heavily on complex diagrams or large external tables requiring multimodal input.

We standardize the selected problems using a structured pipeline that combines LLM-based processing with human oversight. This ensures language clarity, formatting consistency, and reduced risk of data contamination. The pipeline includes the following steps:

\begin{enumerate}
    \item \textbf{Language Normalization:} Non-English problems are translated into fluent English using machine translation, while preserving the original engineering semantics.

    \item \textbf{Expression Rewriting:} To minimize potential overlap with pretraining data, each problem is paraphrased by the LLM using diverse sentence structures and reasoning styles. While surface expressions are significantly altered, the core logic, numerical values, and solution paths remain unchanged. This step produces the \textit{perturbed version} of each task, which is used to evaluate model robustness to superficial input variations.

    \item \textbf{Multimodal Simplification:} For problems containing simple figures or tables, we extract and describe the essential information using plain text or \LaTeX-formatted representations to support uniform text-based evaluation.
\end{enumerate}

\textbf{LLM Prompt Template:} The following instruction prompt is used to guide the LLM in modifying each problem:
    
    \begin{lstlisting}
    """Assuming you are a question expert, please translate this question into English. And while ensuring that the meaning of the question remains unchanged (preserving all logic, values, and the type of reasoning required), change the way the question is expressed by rewriting it in a way that is radically different from your regular logical structure, simulating the randomness of manual rewriting by human experts, and using as many sentence variations as possible. If there is a table, please convert it into a table form using LaTeX. For simple pictures, please describe them directly. The question is required to be converted into is in str format."""
    \end{lstlisting}
To ensure the technical rigor and domain consistency of the Level 3 dataset, the entire generation and transformation process was closely supervised and iteratively revised by doctoral-level professionals with extensive expertise in engineering and mathematical modeling. These experts reviewed both the selection of source problems and the outputs produced by the language model, verifying that each task preserved the original problem’s intent, accurately reflected real-world engineering reasoning, and met the standards expected in academic and professional modeling contexts.

The details of how the original contest scoring standards were mapped into EngiBench’s formal scoring rubrics are described in the later subsection (see Section \ref{sec: Level 3 Evaluation Details}).

\subsection{Version Variant Generation}
To assess model robustness and isolate specific reasoning limitations, we generate three structured variants for each Level 1 and Level 2 problem: \textit{Perturbed}, \textit{Knowledge-Enhanced}, and \textit{Math Abstraction}. These variants are created through LLM prompting, with manually verified outputs to ensure alignment with the original problem logic and correctness. Below, we describe the purpose and generation criteria for each variant, accompanied by illustrative prompts.

\begin{itemize}
    \item \textbf{Perturbed Variant.}  
    This variant alters the surface form of the original problem—either through numerical or linguistic changes—while preserving its core logic and computational requirements. The purpose is to test whether model performance stems from true reasoning ability or superficial pattern matching. A rewriting suitability code (0–3) guides the type of modification to apply. The prompt used to generate the perturbed version and related content is as follows:

    \begin{lstlisting}
"""
1. Rewriting Suitability: Determine the type (0-3):
   - 0: Non-rewritable (use only when necessary)
   - 1: Modify expressions only
   - 2: Modify numerical values only
   - 3: Modify both expressions and numerical values
   // Note: All rewrites must maintain the original problem logic, engineering context, and reasoning/computational requirements

2. Rewritten Problem: Rewrite the problem according to the type of rewriting suitability above. Make the answer as difficult as possible while ensuring that the answer is correct. (Please rewrite the problem in a way that is radically different from your regular logical structure by: (1) avoiding common reasoning patterns in your model, (2) simulating human expert manual rewriting randomness, and (3) using maximum sentence variation.)
   - If 0, return original problem unchanged
   - If 1, modify expressions only
   - If 2, modify numerical values only
   - If 3, modify both expressions and values

3. Rewritten Solution Process: Provide step-by-step explanation including all reasoning, calculations and logic. Clearly state if answer can be obtained directly through formula substitution (shortest solution path without intermediate steps).

4. Rewritten Answer: Provide correct answer for rewritten problem (only types 2/3 may change)"""
    \end{lstlisting}

    \item \textbf{Knowledge-enhanced Variant.}  
    In this version, relevant domain knowledge—such as formulas, constants, and conversions—is explicitly provided before the original question. This allows us to evaluate whether performance deficits are due to missing knowledge or failures in application. The question itself is unchanged to isolate the impact of added context. The prompt used to generate the knowledge-enhanced version is as follows:

    \begin{lstlisting}
"""Knowledge-Enhanced Version:
WARNING: Make sure the final numerical answer to the converted mathematical problem is exactly the same as the original problem.

Given:
- List all relevant formulas or principles (e.g., Ohm's Law: V = I * R)
- Include physical constants with values if they are involved (e.g., g = 9.8 m/s^2)
- Specify unit conversions if applicable (e.g., 1 kWh = 3.6 * 10^6 J)
- State any assumptions or ideal conditions if necessary (e.g., assume no heat loss)

Problem:
Repeat the original question exactly as stated

Example:
Original: "Calculate voltage across 5 Ohm resistor with 2 A current"
Enhanced:
"Given:
- Ohm's Law: V = I * R
- Problem: Calculate voltage across 5 Ohm resistor with 2 A current" """
    \end{lstlisting}

    \item \textbf{Math Abstraction Variant.}  
    This version reformulates the original engineering problem into a purely mathematical format by removing all domain-specific context. Variables and operations are explicitly defined to preserve the exact calculation logic. This allows us to isolate whether reasoning failure arises from contextual understanding or mathematical ability. The prompt used to generate the math abstraction version is as follows:

    \begin{lstlisting}
"""Rewrite the given problem into a purely mathematical version by:

a. Remove all domain-specific context (e.g., chemistry, physics, economics).
b. Keep only numbers, variables, and math operations.
c. If domain-specific knowledge is required (e.g., reaction ratio, atomic mass), extract only the final numerical ratio or constant and include it directly.
d. Maintain the exact calculation logic and final answer.
e. Use structured symbolic language in a compact form:
- Introduce variables explicitly (e.g., "Let x = 2 and y = 3.")
- Define the calculation clearly (e.g., "Total z = min(x, y) * 2.")
- End with "Find the result."

WARNING: Make sure the final numerical answer to the converted mathematical problem is exactly the same as the original problem.

Examples:

Original: "In the reaction: Cl2 + H2 -> 2HCl, 1 mole of Cl2 reacts with 2 moles of H2. How many moles of HCl can be formed?"
converted_problem: "Let x = 1 and y = 2. They react in the ratio x : y : z = 1 : 1 : 2. Total product z = min(x, y) * 2. Find the result."

Original: "A 2m wide platform sinks 0.01m under 60kg. Estimate its length assuming water density = 1000 kg/m^3."
converted_problem: "Let x = 60 / (2 * 0.01 * 1000). Find the result." """
    \end{lstlisting}

\end{itemize}

\section{Dataset URLs, License, and Hosting Plan}

EngiBench is released for research and evaluation purposes only. All third-party artifacts are used in accordance with their original licenses. The released benchmark does not redistribute restricted original content, and commercial use of the benchmark is not permitted.

\subsection{Dataset Instance Metadata}
For the EngiBench dataset, each instance corresponds to an engineering task and is stored in a structured format. Instances are categorized according to task difficulty (Level 1, 2, or 3) and are constructed with multiple versions to enable fine-grained evaluation of different capabilities. The metadata fields for each level are described below:

\paragraph{\textbf{Level 1 and Level 2}}
Each row in the Level 1 \& 2 dataset corresponds to a closed-form or structured engineering problem, and includes the following fields:

\begin{itemize}
  \item \textbf{problem} -- Original natural language problem statement.
  \item \textbf{answer} -- Ground truth answer to the original problem.
  \item \textbf{subfield} -- Engineering subfield to which the problem belongs (e.g., Systems \& Control).
  \item \textbf{category} -- Topic-specific classification within the subfield (e.g., Thermodynamics).
  \item \textbf{difficulty} -- Either \texttt{Level 1} (Foundational Knowledge Retrieval) or \texttt{Level 2} (Contextual Reasoning).
  \item \textbf{converted\_problem} -- Abstract mathematical formulation of the problem.
  \item \textbf{converted\_problem\_llm\_answer} -- LLM-generated response to the converted problem.
  \item \textbf{knowledge\_enhanced\_problem} -- Problem reformulated with explicit formulas and domain definitions.
  \item \textbf{rewritten\_problem} -- Semantically or numerically perturbed variant of the original problem.
  \item \textbf{rewritten\_answer} -- Answer to the rewritten problem.
  \item \textbf{rewritten\_converted\_problem} -- Mathematical abstraction of the rewritten problem.
  \item \textbf{rewritten\_converted\_problem\_llm\_answer} -- LLM response to the rewritten converted problem.
  \item \textbf{rewritten\_knowledge\_enhanced\_problem} -- Knowledge-enhanced version of the rewritten problem.
\end{itemize}

\paragraph{\textbf{Level 3}}
Each Level 3 instance represents an open-ended modeling task and includes both the problem prompt and a rubric-based evaluation across multiple capability dimensions:

\begin{itemize}
  \item \textbf{question} -- English translation of the open-ended modeling task.
  \item \textbf{question\_modified} -- Semantically perturbed variant of the task.
  \item \textbf{source\_detail} -- Source of the modeling task (e.g., MCM, coursework).
  \item \textbf{official\_scoring\_standard} -- English translation of rubric criteria.
  \item \textbf{subfield} -- Engineering subfield of the task.
  \item \textbf{category} -- Domain or topic under which the task is categorized.
  \item \textbf{information\_extraction\_score} -- Score for identifying relevant variables and constraints.
  \item \textbf{multi\_objective\_decision\_score} -- Score for resolving trade-offs across objectives.
  \item \textbf{uncertainty\_handling\_score} -- Score for reasoning under ambiguity or variable inputs.
  \item \textbf{domain\_specific\_reasoning\_score} -- Score for applying engineering-specific logic and formulas.
\end{itemize}

\section{Evaluation Details}

\subsection{Level 1 \& Level 2 Evaluation Details}\label{sec: Level 1 Level 2 Evaluation Details}

Level~1 and Level~2 tasks consist of well-defined problems with clearly defined and unique solutions. We therefore adopt a \textit{binary scoring} scheme, where each model-generated answer is compared against a reference answer and marked as either correct (1) or incorrect (0). Overall performance is reported in terms of accuracy.

Evaluation is conducted through an automated comparison procedure. To handle diverse numerical formats, units, and equivalent expressions, we design a standardized evaluation prompt, which is independently executed by GPT-4.1 and Gemini~2.5~Flash. For cases where the two evaluators produce inconsistent judgments, manual verification is performed to determine the final decision. The evaluator determines whether a generated answer matches the reference answer based on mathematical correctness, unit validity, and logical consistency. For numerical questions, a tolerance of $\pm$2\% is allowed to account for rounding effects in multi-step calculations. The evaluator is instructed to output only a Boolean decision (``True'' or ``False'') to ensure consistent and reproducible scoring.

To verify evaluation consistency and reliability, we perform multi-model cross-checking and human spot checks. Specifically, all Level~1--2 responses are independently evaluated by GPT-4.1 and Gemini~2.5~Flash, and their results are compared. In addition, we randomly sample 300 problems for manual verification. On this subset, GPT-4.1 achieves an evaluation accuracy of 98.67\%, while Gemini~2.5~Flash achieves 98.33\%. These results demonstrate consistent evaluation behavior and show that the automated procedure closely aligns with deterministic answer matching for Level~1 and Level~2 tasks.

\begin{lstlisting}
"""Please analyze these two answers carefully:
Generated Answer: {generated_answer}
Standard Answer: {correct_answer}

Follow these rules for comparison:
1. For calculation-focused problems:
   - If the numerical values match, consider it correct even if units are missing
   - Focus on the mathematical reasoning and final numerical result
   - Check if the core calculation steps are correct
   - For complex calculations, allow ±2% tolerance in the final numerical result

2. For conceptual or unit-specific problems:
   - Units and their consistency must be considered
   - The complete answer including units is required

3. Consider the answer correct if:
   - The mathematical reasoning is sound
   - The final numerical value matches (within ±2% tolerance for complex calculations)
   - For calculation-focused problems, matching units are not mandatory

Reply only with "True" or "False". """
\end{lstlisting}

\subsection{Level 3 Evaluation Details}\label{sec: Level 3 Evaluation Details}

\subsubsection{Rubric Construction}\label{sec: Rubric Construction}

To enable systematic evaluation of open-ended modeling tasks, we construct structured scoring rubrics from official contest-provided scoring standards.

For each problem, the official scoring description is paired with the problem statement and provided to an LLM using a carefully designed instruction prompt to generate an initial rubric draft aligned with the four target engineering capabilities. The LLM serves only as an auxiliary tool for structuring and organizing the rubric.

Each rubric is then independently reviewed and cross-checked by two reviewers with engineering backgrounds. In cases of disagreement, final decisions are adjudicated by experts who have won national or international first prizes in engineering modeling competitions, ensuring technical rigor and accuracy.

The prompt used for rubric generation is provided below.

\begin{lstlisting}
"""Assume you are an expert in problem design and grading, with deep familiarity with mathematical modeling. Please help me design an evaluation rubric for assessing large language models' engineering capabilities. Specifically, I will provide a problem and its scoring criteria, and you will tell me which of the following capabilities are assessed by this rubric: redundant_information_filtering_score, multi_objective_tradeoff_score, uncertainty_handling_score, and deep_knowledge_integration_score. In particular, please identify how each capability is assessed through specific aspects of the problem or rubric.

For each capability that is covered, provide a scoring rubric in the following format:

Problem [(Problem ID)]:
redundant_information_filtering_score: (1)(2)...
multi_objective_tradeoff_score: (1)(2)...
uncertainty_handling_score: (1)(2)...
deep_knowledge_integration_score: (1)(2)...

Notes: Each capability has a total possible score of 10 points. In other words, the total score for each listed capability should sum to 10 points. Capabilities that are not covered in this problem receive 0 points. The rubric should further specify, under each capability, the different score levels (e.g., 1 point, 2 points, 3 points, etc.) and the corresponding specific behaviors or response characteristics associated with each level.

Please read the problem and rubric carefully and provide a capability-based evaluation rubric for how this problem assesses the output of large language models."""
\end{lstlisting}

\subsubsection{Rubric-based Scoring and Human Calibration}\label{sec: Rubric-based Scoring and Human Calibration}

The finalized rubrics are applied to evaluate model-generated responses for Level~3 tasks. We implement an automated LLM-based scoring pipeline that assesses solution quality along multiple capability dimensions defined by the rubrics. Specifically, scores are independently produced by GPT-4.1 and Gemini~2.5~Flash, and the final score is obtained by averaging the two to reduce evaluator-specific variability.

To ensure the reliability of the reported results, LLM-generated scores are reviewed and calibrated by annotators with engineering backgrounds. The main results in this paper report calibrated scores. We note that fully automated LLM-based scoring already provides a strong and practical reference, as further supported by the consistency and validity analysis in Appendix~\ref{sec: Level 3 Scoring Consistency Analysis}.

The prompt used to evaluate the generated answer against the rubric is as follows:
\begin{lstlisting}
f"""
    You are a professional modeling competition judge with extensive experience in evaluating mathematical and engineering models. Please conduct a rigorous evaluation of the following answer based on the provided criteria.

    Answer to evaluate:
    {answer}

    Evaluation Criteria:
    {score_criteria}

    Please evaluate strictly according to the criteria and provide your assessment in the following JSON format:
    {{
        "score": <score between 0-10, can use decimal points for precision>,
        "reason": "Detailed evaluation breakdown:\n
                  1. [Specific criterion] - [sub-score] points: [justification]\n
                  2. [Specific criterion] - [sub-score] points: [justification]\n
                  3. [Specific criterion] - [sub-score] points: [justification]\n
                  Final score: [total] points"
    }}

    Note: 
    - Break down your scoring into specific components
    - Provide clear justification for each sub-score
    - Be objective and consistent in your evaluation
    - Consider both the technical accuracy and the methodology
    """
\end{lstlisting}

\subsection{Level 3 Scoring Examples}

As results shown in section \ref{sec: Performance for Level 3 Tasks}, the answers of LLMs to open-ended tasks show significant differences in four dimensions of information extraction, multi-objective decision making, uncertainty handling and domain-specific reasoning. Figure \ref{fig:llama4} preliminarily presents two scoring segments, 3 points and 8 points, for the evaluation of models' answers. To demonstrate the response performance of different segments more clearly and intuitively, we provide the following examples with more Level 3 scoring details:

\begin{enumerate}

    \item \textbf{Full Mark (Avg. Score: 9.475):} The problem requires optimizing Hu sheep farm pen utilization under stochastic conditions (conception rates, gestation periods, litter sizes) while adhering to strict capacity constraints and cohabitation rules. The solution must minimize expected losses from idle pens (1 unit/day) or shortages (3 units/day) through dynamic scheduling and statistical validation.
    \begin{itemize}
        \item \textbf{Information Extraction (10/10):} 
        
        Exclusion of Deterministic Assumptions (5/5): 
        Section 1 (System Overview) clarifies all critical parameters modeled as random variables (e.g., ``\( X_c \sim \text{Binomial}(N_m, 0.85) \): Number of successful conceptions; \( G \sim U[147,150] \): Gestation days; \( L_s \sim \text{Poisson}(\lambda = 2.2) \): Liveborn lambs per ewe, with 3\% mortality (\( L_a = L_s \cdot 0.97 \)); \( L_d \sim U[35,45] \): Lactation days'').
        Section 3A (Scenario Generation) replaces fixed values with dynamic sampling (e.g., ``For each scenario, sample: - Which ewes conceive (Bernoulli, 85\%) - Their gestation (\( G \)) - Number of lambs (\( L_s \)), apply mortality - Lactation length (\( L_d \))'').
        Section 6B (Robust Planning) makes flexible scheduling responsive to stochastic outcomes (e.g., ``Adjust mating/rest period within allowed windows to shift animal flows.'').

        Identification of Valid Uncertainty Parameters (5/5):
        Section 1 clarifies explicit distributions for all uncertainties (e.g., ``\( X_c \sim \text{Binomial}(N_m, 0.85) \)... \( G \sim U[147,150] \)... \( L_s \sim \text{Poisson}(2.2) \)... \( L_d \sim U[35,45] \)'').
        Section 3A ensures consistent application in scenario generation (e.g., ``Sample conception (Bernoulli), gestation (\( G \)), litter size (\( L_s \)), lactation (\( L_d \)).'').
        Section 5 (Loss Function) offers loss calculation integrating stochastic inputs (e.g., ``\(\mathbb{E}_{scenario} \left[ \sum_{t} [I_t + 3S_t] \right]\)'').
        \item \textbf{Multi-objective Decision making (9.2/10):}

        Minimized Expected Loss \& Output Maximization (4.5/5): 
        Section 5 (Loss Function) contains rigorous mathematical formulation balancing idle (1 unit) vs. shortage (3 unit) costs (e.g., ``Objective: \(\min \mathbb{E}_{scenario} \left[ \sum_{t} [I_t + 3S_t] \right]\) \( I_t = \text{Idle pens}, S_t = \text{Shortages} \)'').
        Section 7B (Robust Planning) includes statistical validation of tradeoffs (e.g., ``Monte Carlo over Scenarios: Simulate losses across all scenarios for each candidate policy.'')
        Section 8 (Results Table) applies quantitative comparison of policies.
        
        Lactation Flexibility \& Fattening Tradeoffs (4.7/5):
        Section 1 (System Overview) makes explicit dynamic linkage between lactation and fattening (e.g., ``\( L_d \sim U[35,45] \): Lactation days → \( F_d = 210 + 2 \cdot (40 - L_d) \): Fattening days'').
        Section 6B (Robust Planning) considers operational use of flexibility to smooth demand (e.g., ``Adjust rest periods to align cohorts, minimizing `loner pens'.'').
        Section 3A (Scenario Generation) has stochastic integration of tradeoff (e.g., ``Sample lactation length (\( L_d \)), impact on fattening (\( F_d \)).'').
        \item \textbf{Uncertainty Handling (9.2/10):}
        
        Stochastic Process Models (4/4):
        Section 1 (System Overview) specifies explicit distributions for all stochastic parameters (e.g., ``\( X_c \sim \text{Binomial}(N_m, 0.85) \), \( G \sim U[147,150] \), \( L_s \sim \text{Poisson}(2.2) \), \( L_d \sim U[35,45] \)'').
        Section 3A (Scenario Generation) implements full Monte Carlo (e.g., ``Generate 1000 scenarios... sample conception (Bernoulli), gestation (\( G \)), litter size (\( L_s \)), lactation (\( L_d \)).'').
        Section 7B (Robust Planning) includes statistical validation of stochastic outcomes (e.g., ``For each candidate policy, simulate losses across all scenarios.'').

        Dynamic Adjustment Strategies (2.7/3):
        Section 1 (Fattening Calculation) establishes mechanistic linkage of lactation-fattening tradeoff (e.g., ``\( F_d = 210 + 2 \cdot (40 - L_d) \): Fattening days adjusted by lactation.'').
        Section 6B (Robust Planning) makes adaptive scheduling but lacks two-way feedback (e.g., ``Adjust rest periods to align cohorts... weekly rolling re-optimization.'').

        Contingency Sets (2.5/3):
        Section 2 (Cohabitation Rules) contains hard-coded tolerance for uncertainty (e.g., ``Group into largest feasible penfuls within 7-day windows.'').
        Section 8 (Statistical Assessment) analyzes multi-scenario sensitivity (e.g., ``Tabulate average loss, shortage probability, and max pen use.'').
        \item \textbf{Domain-specific Reasoning (9.5/10):}

        Integration of Empirical Rules (4/4):
        Section 2 (Cohabitation Rules) adds hard-codes industry constraints into algorithms (e.g., ``7-day tolerance window for nursing ewes, lambs, and resting ewes...  Group into largest feasible penfuls (≤14 fattening lambs/pen, ≤6 nursing ewes/pen).'').
        Section 1 (System Overview) uses embeds empirical flexibility ranges as distributions (e.g., ``\( L_d \sim U[35,45] \): Lactation days... \( R \sim U[18,22] \): Adjustable rest period.'')
        Section 6B (Robust Planning) operationalizes flexible rest rules (e.g., ``Extend rest periods to align cohorts if pens would otherwise idle.'').
        
        Expected Loss Functions (3/3):
        Section 5 (Loss Function) has rigorous probabilistic loss aggregation (e.g., ``\(\min \mathbb{E}_{scenario} \left[ \sum_{t} [I_t + 3S_t] \right]\) \( I_t = \max(P_{avail} - P_{req}(t), 0) \), \( S_t = \max(P_{req}(t) - P_{avail}, 0) \).'').
        Section 8 (Results Table) quantifies loss distribution across scenarios.
        Section 3B (State Evolution) links stochastic occupancy to loss calculation (e.g., ``For each day \( t \): Compute \( P_{req}(t) \) from sampled cohorts.'').
        
        Stochastic Optimization Algorithms (2.5/3):
        Section 7B (Robust Planning) applies sample average approximation (SAA) method (e.g., ``Monte Carlo simulation over 1000 scenarios to evaluate policies.'').
        Section 6A (Rolling Horizon) uses heuristic dynamic programming (e.g., ``Re-optimize mating batches weekly to maximize cohabitation.'').
        
    \end{itemize}

    \item \textbf{5 points (Avg. Score: 5.375):} The problem involves modeling a team coordination exercise (``Unity Drum'') where 8 members control a drum's tilt by pulling ropes to bounce a ball. Key tasks include: 1. Calculating the drum's tilt angle at t=0.1s based on force/timing inputs (Table 1), accounting for initial 11cm displacement. 2. Ensuring physics-based accuracy in torque, angular acceleration, and geometric relationships.
    \begin{itemize}
        \item \textbf{Information Extraction (7.5/10):} 

        Error Source Analysis (5/6):
        Explicit Recognition: Timing errors-``Some members may apply force slightly before others'' (Algorithm section); strength variation-``Members likely have different strengths'' (Considerations). Partial Implementation: Timing logic in code (if \(\text{timing}[i] \leq 0.1\)) is noted but lacks vector-time coupling; force scaling (\(\text{effective\_force} = \frac{\text{force}(\text{member\_id} - 1)}{10}\)) is arbitrary.
        
        Physical Model Simplification (2.5/4):
        Justified Simplifications: ``Ignores damping for short-duration calculation'' (Considerations); Drum as uniform cylinder (\(I = 0.5 \cdot \text{drum\_mass} \cdot r^2\)). Over-Simplifications: Fixed torque angle (\(\sin\left(\frac{\pi}{2}\right)\)) ignores vector geometry; rope tautness assumption (``If the drum tilts too far, ropes could slack'') not modeled. 
        \item \textbf{Multi-objective Decision making (6.5/10):}

        Tilt Angle and Force Relationship (4.5/6):
        Physics Foundation: Correctly derives torque (\(\tau = r \cdot F \cdot \sin(\theta)\)), inertia (\(I = 0.5 \cdot m \cdot r^2\)), and angular kinematics (\(\theta = \theta_0 + \frac{1}{2} \alpha t^2\)); maps rope geometry (\(\text{angle\_radians} = (\text{member\_id} - 1) \cdot \left( \frac{2\pi}{8} \right)\)). Implementation Gaps: Timing logic (if \(\text{timing}[i] \leq 0.1\)) is crude; forces are binary (on/off) rather than time-interpolated; no optimization for tilt minimization (e.g., predictive control or force balancing).

        Computational Efficiency (2/4):
        Basic Looping-iterates over 8 members with O(1) operations per member (e.g., \(\text{torque} = \text{drum\_radius} \cdot \text{force} \cdot \sin\left( \frac{\pi}{2} \right)\)). No Advanced Techniques-lacks vectorization, memoization, or scalability for larger teams.
        \item \textbf{Uncertainty Handling (2/10):}

        Error Propagation Analysis (2/4):
        Acknowledgment Only: Mentions ``members likely have different strengths and reaction times'' (Considerations); suggests ``extended to simulate more realistic distributions'' but provides no math or implementation. No Quantification: Lacks sensitivity analysis or error bounds on tilt angle.
        
        Numerical Simulation \& Estimation (0/4):
        No Monte Carlo: Code calculates tilt for fixed inputs only (\text{force\_data}); no randomization of force/timing or statistical output (mean/variance).
        
        Methodological Clarity (N/A): Physics steps are clear but irrelevant to uncertainty scoring.
        \item \textbf{Domain-specific Reasoning(5.5/10):}

        3D Mechanics Modeling (2.5/6):
        2D Limitation: Explicitly states ``our coordinate system will be planar (X and Y only)'' (Key Equations); torque calculation (\(\tau = r \cdot F \cdot \sin(\theta)\)) ignores out-of-plane forces. Partial Physics: Correctly models drum as cylinder (\(I = 0.5 \cdot m \cdot r^2\)) but lacks 3D rotation dynamics.
        
        Model-Based Optimization Strategy (3/4):
        Suggestions Without Implementation: Proposes ``damping term proportional to angular velocity'' (Considerations); mentions ``member variation'' but no adaptive control (e.g., PID for tilt correction).
    \end{itemize}

    \item \textbf{1 point (Avg. Score: 1.25):} The problem involves coordinating multiple meteorological units (each with 1 primary and 2 secondary stations) to ensure reliable hourly weather data collection and full data sharing under strict communication constraints. Key challenges include managing transmission reliability (80\% for secondaries, 100\% for primaries), message capacity limits, and achieving ≥97\% success probability within 8 minutes for primary data exchange. The goal is to determine the maximum number of units (Nmax), design transmission schemes, and compute performance metrics.
    \begin{itemize}
        \item \textbf{Information Extraction (2/10):} 
        High-Probability Constraint Processing (0/5):
        Failure to Address Probabilistic Guarantee: The answer calculates secondary transmission success as ``expected number of reports received... is \(4 \times 0.8 = 3.2\)'' (Step 4) but never models retransmissions or redundancy to achieve ≥97\% success. The assumption of direct success ignores the problem's explicit probability requirement. Missing Critical Logic: No discussion of how to compensate for the 20\% failure rate (e.g., retrying failed transmissions, acknowledgments, or error correction).
        
        Time Window Isolation (2/5):
        Interleaved Logs Without Justification: The primary and secondary transmission logs (Tables 1 \& 2) are interleaved in the solution (``Round 1: Primary 1→2; Round 1: Secondary 1→1a''), but no protocol ensures collision avoidance (e.g., TDMA, priority scheduling). Unverified Simultaneity Assumption: The answer states ``Simultaneous reception allowed during transmission'' (Step 1) but doesn't prove this suffices for concurrent primary/secondary transmissions under the 8-minute constraint.
        \item \textbf{Multi-objective Decision making (2/10):}

        3D Parameter Optimization (0/6):
        Single-Parameter Focus: The answer only optimizes for \textbf{N\_max} (``\(N(N-1)/2 ≤ 8\) → \(N_{\text{max}} = 4\)'', Step 2) but ignores joint optimization of capability (no analysis of 158-character message limits or segment splitting efficiency), reliability (no adjustment for secondary station 80\% success rate such as no retransmission strategy) and time (assumes 8 minutes suffice without validating secondary transmission overhead). Missed Pareto Frontier: Fails to explore tradeoffs (e.g., ``Could N=5 work if secondary transmissions are reduced?'').
        
        Resource Allocation Strategy (2/4):
        Equal Bandwidth Only: Primary stations follow a round-robin schedule (``1→2, 1→3, 1→4, 2→3, ...'', Table 1), and secondaries transmit uniformly (``1→1a, 1→1b, 2→2a, ...'', Table 2). No Prioritization: Critical objectives (e.g., ensuring ≥97\% success) aren't prioritized in scheduling. 
        \item \textbf{Uncertainty Handling (0/10):}

        High-Order Probability Events (0/6):
        No Threshold Calculation: The answer states secondary stations have an ``80\% transmission/reception success rate'' (Step 1) but never computes the probability of achieving ≥97\% success (e.g., via binomial distribution for multiple retries). Misleading Metric: The ``mean secondary reports received per primary station (3.2)'' (Step 4) is irrelevant to the cumulative success probability requirement.
        
        Asymmetric Loss (0/4):
        No Cost Analysis: The solution ignores idle time cost (unused transmission slots due to failures) and rental loss (penalties for delayed data delivery implied by "critical rescue operations").
        \item \textbf{Domain-specific Reasoning (1/10):} 

        Mixed-Integer Programming (0/5):
        No Optimization Model: The answer derives \(N_{\text{max}} = 4\) via a simple inequality (``\( \frac{N(N - 1)}{2} \leq 8 \)'', Step 2) but lacks an objective function (e.g., ``maximize N while meeting time/reliability constraints''), and omits integer constraints (N must be discrete) or linear relaxation techniques. Ad-Hoc Calculation: No use of MINLP (Mixed-Integer Nonlinear Programming) to jointly optimize N, transmission scheduling, and reliability.
        
        Fault-Tolerant Protocol Design (1/5):
        Basic Segmentation: Mentions ``reports can split into two 50-character segments'' (Step 1) but no dual verification (never states if segments are sent redundantly to different primaries) and no formal protocol (assumes secondary stations report to all primaries without fault recovery like checksums, ACKs).
    \end{itemize}
    
\end{enumerate}

\section{Additional Analysis}

\subsection{Level 3 Scoring Consistency Analysis}\label{sec: Level 3 Scoring Consistency Analysis}

\begin{table*}[t]
\centering
\caption{Level~3 average scores under three scoring variants. Human-calibrated scores are reported as the main results; LLM-only scores are produced by the automated scoring pipeline; Human-only scores are reference scores from official solutions and expert grading.}
\label{tab:level3_average_three_scoring_variants}
\setlength{\tabcolsep}{10pt}
\renewcommand{\arraystretch}{1.15}
\begin{tabular}{l cc cc cc}
\toprule
\multirow{2}{*}{Model} &
\multicolumn{2}{c}{Human-calibrated} &
\multicolumn{2}{c}{LLM-only} &
\multicolumn{2}{c}{Human-only} \\
\cmidrule(lr){2-3}\cmidrule(lr){4-5}\cmidrule(lr){6-7}
& Original & Perturbed & Original & Perturbed & Original & Perturbed \\
\midrule
Human Expert        & 8.728 & 8.736 & 8.697 & 8.702 & 8.735 & 8.729 \\
GPT-4.1             & 7.108 & 7.002 & 7.053 & 6.972 & 7.208 & 7.043 \\
Claude 3.7 Sonnet   & 6.656 & 6.445 & 6.713 & 6.619 & 6.970 & 6.526 \\
GPT-4.1 Mini        & 6.793 & 6.378 & 6.581 & 6.334 & 6.705 & 6.558 \\
DeepSeek-V3         & 6.289 & 6.329 & 6.358 & 6.264 & 6.396 & 6.386 \\
Gemini 2.5 Flash    & 6.197 & 6.208 & 6.002 & 6.145 & 6.063 & 6.185 \\
Gemini 2.0 Flash    & 6.145 & 5.991 & 5.989 & 5.902 & 6.167 & 6.035 \\
GPT-4.1 Nano        & 5.968 & 5.738 & 5.764 & 5.673 & 6.074 & 5.882 \\
GLM-4-32B           & 5.615 & 5.653 & 5.860 & 5.761 & 5.760 & 5.694 \\
GLM-4-9B            & 4.833 & 5.169 & 5.079 & 5.227 & 4.822 & 5.168 \\
Claude 3.5 Sonnet   & 5.228 & 5.049 & 5.317 & 5.254 & 5.187 & 5.106 \\
Llama 3.3           & 4.837 & 5.008 & 4.937 & 4.804 & 4.939 & 5.055 \\
Qwen2.5-72B         & 4.935 & 4.722 & 4.836 & 4.665 & 5.007 & 4.831 \\
Qwen2.5-7B          & 4.339 & 4.619 & 4.580 & 4.591 & 4.362 & 4.669 \\
Llama 4             & 3.767 & 3.861 & 3.808 & 3.926 & 3.943 & 3.892 \\
DeepSeek-R1 7B      & 4.043 & 3.810 & 3.775 & 3.648 & 4.105 & 3.989 \\
Mixtral-8$\times$7B & 3.203 & 3.476 & 3.110 & 3.279 & 3.372 & 3.577 \\
\bottomrule
\end{tabular}
\vspace{-1mm}
\end{table*}

To further examine the reliability of our evaluation protocol, we compare three scoring variants for Level~3 tasks: human-calibrated scoring (reported as the main results), fully automated LLM-only scoring, and fully manual human scoring (human-only).

Table~\ref{tab:level3_average_three_scoring_variants} summarizes the average scores under these three scoring settings for both original and perturbed tasks. Across models and task settings, LLM-only scores are generally close to human-calibrated scores, with differences typically within a small margin. This indicates that the proposed rubric enables reliable automated evaluation, while human calibration mainly serves to correct a limited number of edge cases and ensure maximum rigor in the reported results.

Regarding the validation of scoring consistency, the relative ordering of models remains largely consistent across the three scoring variants. Notably, the human-only scoring serves as a ground truth baseline, confirming that the trends observed in automated and calibrated scoring are robust.

These results support the practical use of fully automated scoring for large-scale benchmarking, while human calibration provides additional assurance when reporting final evaluation results.

\begin{figure*}[t]
\centering
\captionsetup{font={small}, skip=8pt}
\includegraphics[width=0.9\textwidth]
{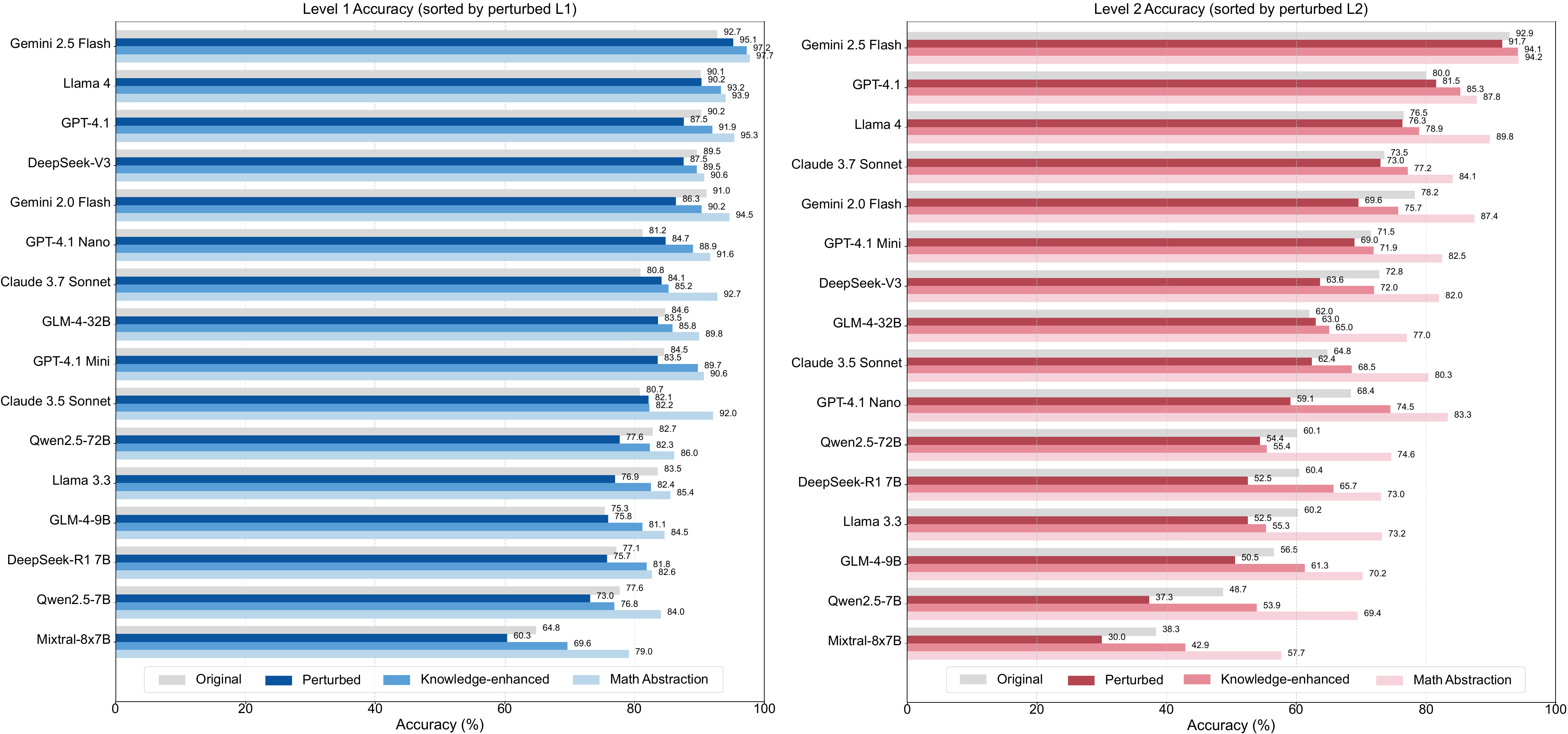}
\caption{Accuracy of LLMs on Level 1 (left) and Level 2 (right) tasks across four variants: Original, Perturbed, Knowledge-enhanced, and Math Abstraction. Drops in the Perturbed version indicate sensitivity to input changes, while gains in the latter two show that current LLMs require external knowledge or reformulation to improve accuracy—highlighting their lack of these abilities.
}
\label{appfig: level1&2}
\vspace{-6mm}
\end{figure*}

\subsection{Level 1 Analysis}

\textbf{Minor perturbations cause performance drops, revealing shallow generalization.}\quad
Figure~\ref{appfig: level1&2} (left) presents model accuracy on Level 1 tasks across four input variants: Original, Perturbed, Knowledge-enhanced, and Math Abstraction. When problems are perturbed through minor changes in wording or numerical values, average model accuracy drops from 82.9\% to 81.5\%. Notably, Llama 3.3 and Qwen2.5-72B decline by 6.6\% and 5.1\%, respectively. This indicates that some models exhibit limited robustness and often rely on memorized phrasing or surface patterns rather than generalizable reasoning.

\textbf{Explicit knowledge prompts mitigate reasoning failures in weaker models.}\quad
When explicit domain knowledge—such as formulas, constants, or unit conversions—is added to the input, accuracy improves to 85.5\% on average. Weaker models benefit the most: GPT-4.1 Mini gains 6.2\% and Mixtral-8x7B improves by 9.3\%. This pattern suggests that many errors are not caused by a complete lack of knowledge, but rather by the inability to retrieve and apply relevant concepts without targeted prompting. Explicitly embedding domain knowledge thus serves as an effective intervention for enhancing reasoning activation.

\textbf{Removing contextual language highlights semantic limitations.}\quad
Performance further increases to 89.4\% when problems are rewritten into abstract mathematical form, removing all contextual language. For example, Qwen2.5-7B and Mixtral-8x7B improve by 10.9\% and 18.8\%, respectively. This reveals that most Level 1 failures are not due to weak computational ability, but rather arise during semantic interpretation and variable binding. Once language ambiguity is removed, models can more reliably execute the required calculations, underscoring a gap between symbolic proficiency and contextual understanding.

\subsection{Level 2 Analysis}

Level 2 tasks emphasize multi-step reasoning under structured constraints, making them more sensitive to input variability. As shown in Figure~\ref{appfig: level1&2} (right), the average model accuracy declines from 66.6\% on the Original version to 61.6\% on the Perturbed variant. This 5.0\% drop indicates that even minor changes to semantic phrasing or numerical values can significantly disrupt reasoning chains. For instance, GPT-4.1 Nano drops by 9.3\% and Qwen2.5-7B by 11.4\%, revealing their limited robustness when facing contextual and structural perturbations in problem inputs.

Incorporating explicit domain knowledge helps reduce ambiguity and recover performance. With knowledge-enhanced inputs, the average accuracy rises to 68.6\%, a 7.0\% improvement over the perturbed baseline. Larger gains are observed for models such as GPT-4.1 Nano (+15.4\%) and Qwen2.5-7B (+16.6\%), suggesting that knowledge prompts assist in constraint interpretation and formula selection. However, some models such as DeepSeek-V3 show minimal improvement, implying that knowledge access alone may not compensate for limitations in multi-step reasoning capabilities.

Symbolic abstraction of Level 2 tasks into pure mathematical form results in the largest performance gains. The average accuracy increases to 79.2\%, with many models gaining over 15\%. This trend is especially prominent for weaker models like Qwen2.5-7B (from 37.3\% to 69.4\%) and Mixtral-8x7B (from 30.0\% to 57.7\%). These improvements confirm that many model failures stem not from computational weakness, but from difficulties parsing, organizing, and executing the reasoning steps embedded in natural language problem statements. This underscores the importance of assessing upstream cognitive processes that precede symbolic computation—dimensions often underexamined in traditional mathematical benchmarks.

\subsection{Level 3 Analysis}
\vspace{-2mm}
\begin{figure*}[t]
  \centering
  \includegraphics[width=0.8\linewidth]{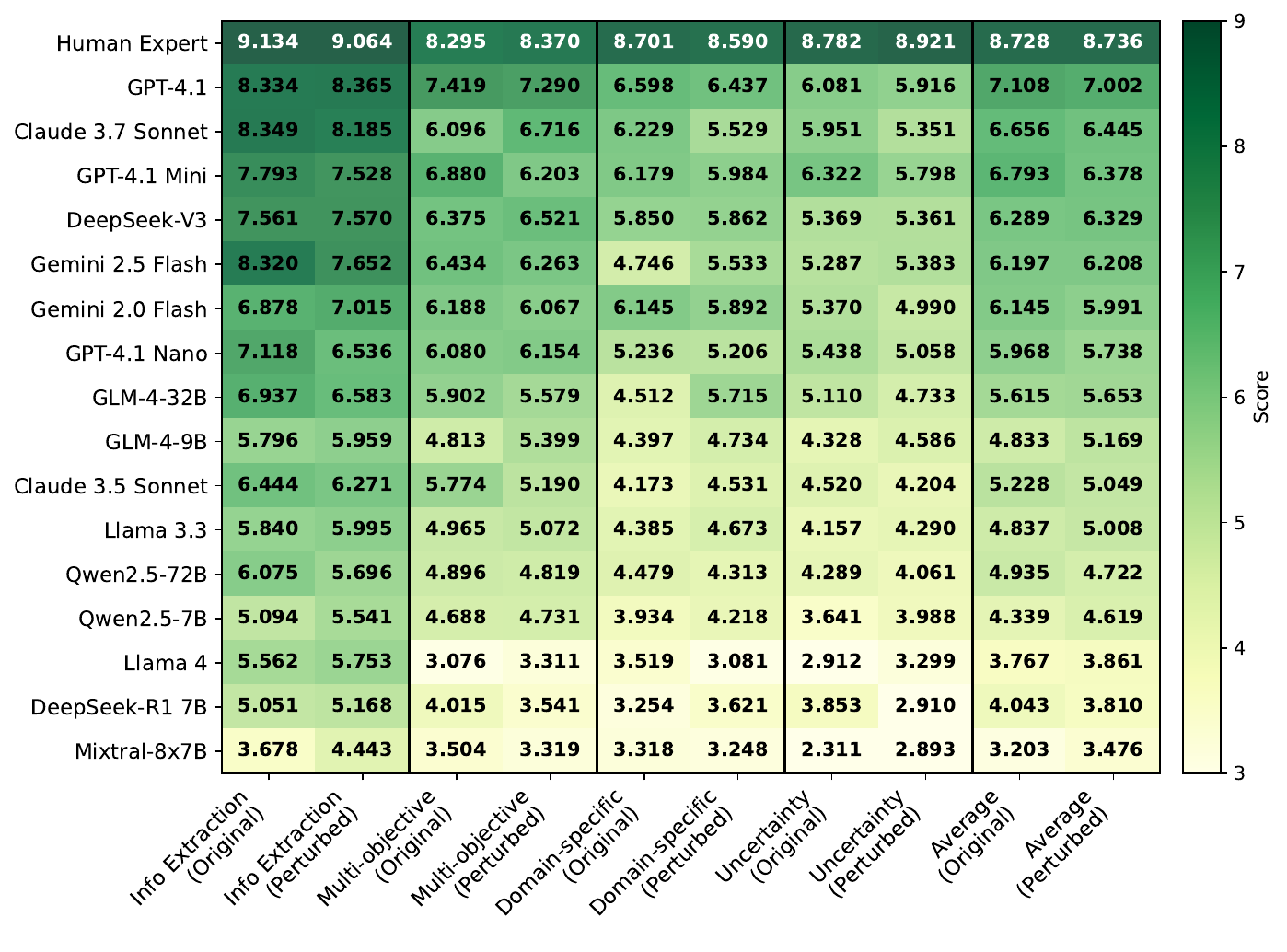}
  \caption{Level 3 Model Evaluation. The figure presents average model performance on Level 3 tasks across four capability dimensions, including information extraction, domain-specific reasoning, multi-objective decision-making, and uncertainty handling, under both original and perturbed problem formulations.}
  \label{fig:level3_only}
\vspace{-2mm}
\end{figure*}

Figure~\ref{fig:level3_only} presents the performance of various models across four key capabilities: Redundant Information, Multi-Objective Decision, Domain Knowledge, and Uncertainty Handling. The results are further separated into \textit{original} and \textit{perturbed} problem formulations. Overall, human experts substantially outperform all models across all dimensions, with average scores of 8.73 (original) and 8.74 (perturbed). In contrast, LLMs demonstrate significantly lower scores, revealing a persistent gap between current LLMs' capabilities and human-level reasoning. The average model scores before and after rewriting are 5.372 and 5.341, respectively—a marginal difference of only 0.58\%. This indicates that most models possess a reasonable degree of generalization, and the benchmark shows no signs of data contamination across reformulated prompts, preserving task consistency. 

Based on the overall average scores, we categorize model performance into three tiers:

\textbf{Tier 1 (Average Score > 6.5)}\quad 
This tier includes GPT-4.1, Claude 3.7 Sonnet, and GPT-4.1 Mini. These models demonstrate strong performance across all four evaluated capabilities. In particular, their scores in Information Extraction and Multi-Objective Decision often exceed 7, approaching human expert levels. Their performance in Domain Knowledge and Uncertainty Handling also remains consistently above 6, indicating robust reasoning capabilities and broad task adaptability.

\textbf{Tier 2 (Average Score $\approx$ 5.5--6.5)}\quad 
This tier consists of DeepSeek-V3, Gemini 2.5 Flash, Gemini 2.0 Flash, GPT-4.1 Nano, and GLM-4-32B. These models achieve reasonable performance in Information Extraction and Multi-Objective Decision, but exhibit noticeable weaknesses in Domain Knowledge and Uncertainty Handling, where scores commonly fall below 6. Some models approach the 5-point threshold in these dimensions, reflecting limitations in complex reasoning and knowledge integration.

\textbf{Tier 3 (Average Score < 5.5)}\quad 
This tier includes GLM-4-9B, Claude 3.5 Sonnet, Llama 3.3, Qwen2.5-72B, Qwen2.5-7B, Llama4, DeepSeek-R1 7B, and Mixtral-8x7B. These models consistently underperform across all four capabilities, typically scoring between 3 and 5. Their weakest areas are Domain Knowledge and Uncertainty Handling, where some models fall below 4. These results indicate substantial deficiencies in background reasoning and generalization to ambiguous or underspecified tasks.

\subsection{Subfield Performance Analysis}

\begin{figure*}[htbp]
    \centering
    \includegraphics[width=1\textwidth]{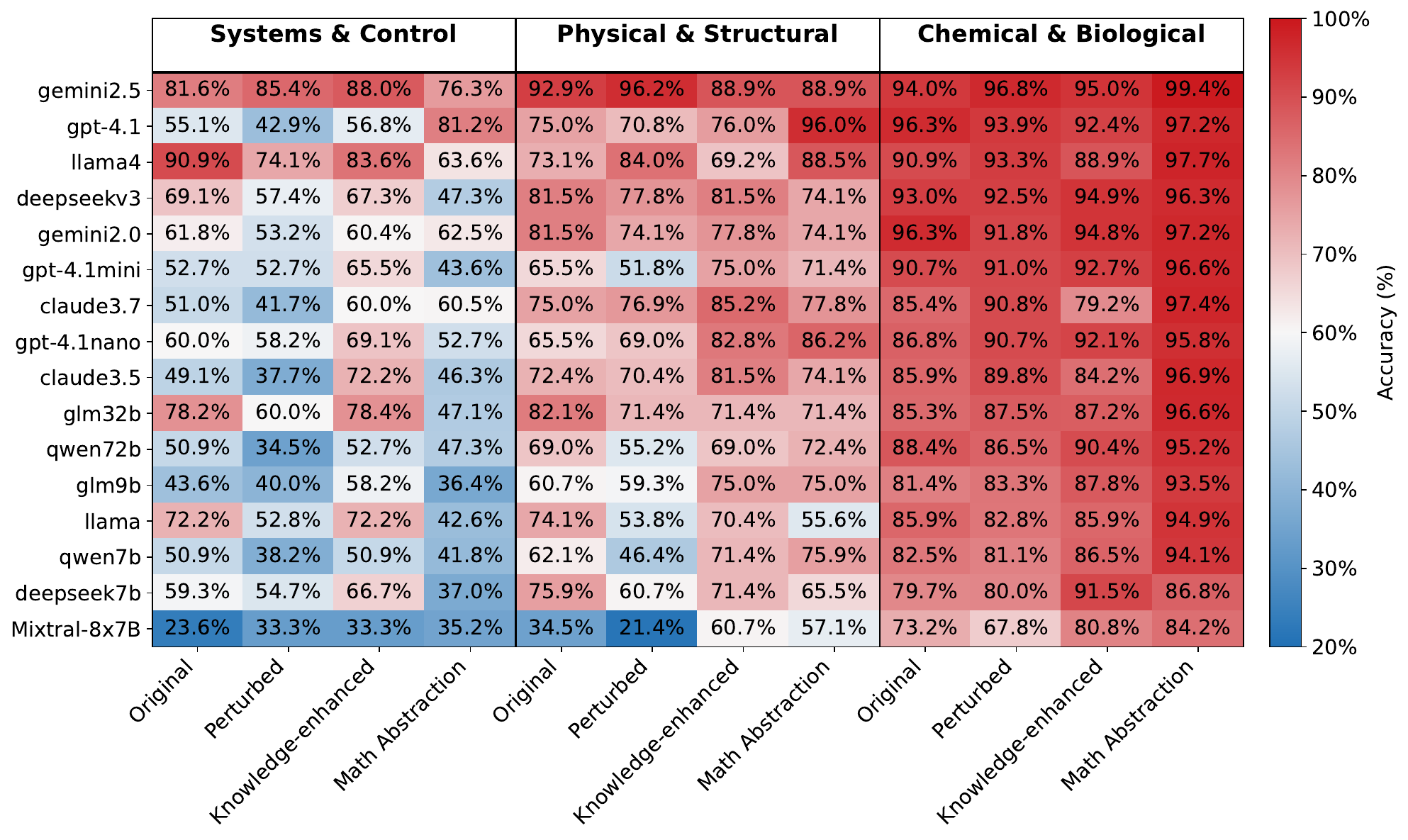}
    \caption{Accuracy across engineering subfields and problem variants in Level~1.}
    \label{fig:level1_heatmap}
\end{figure*}

\begin{figure*}[htbp]
    \centering
    \includegraphics[width=1\textwidth]{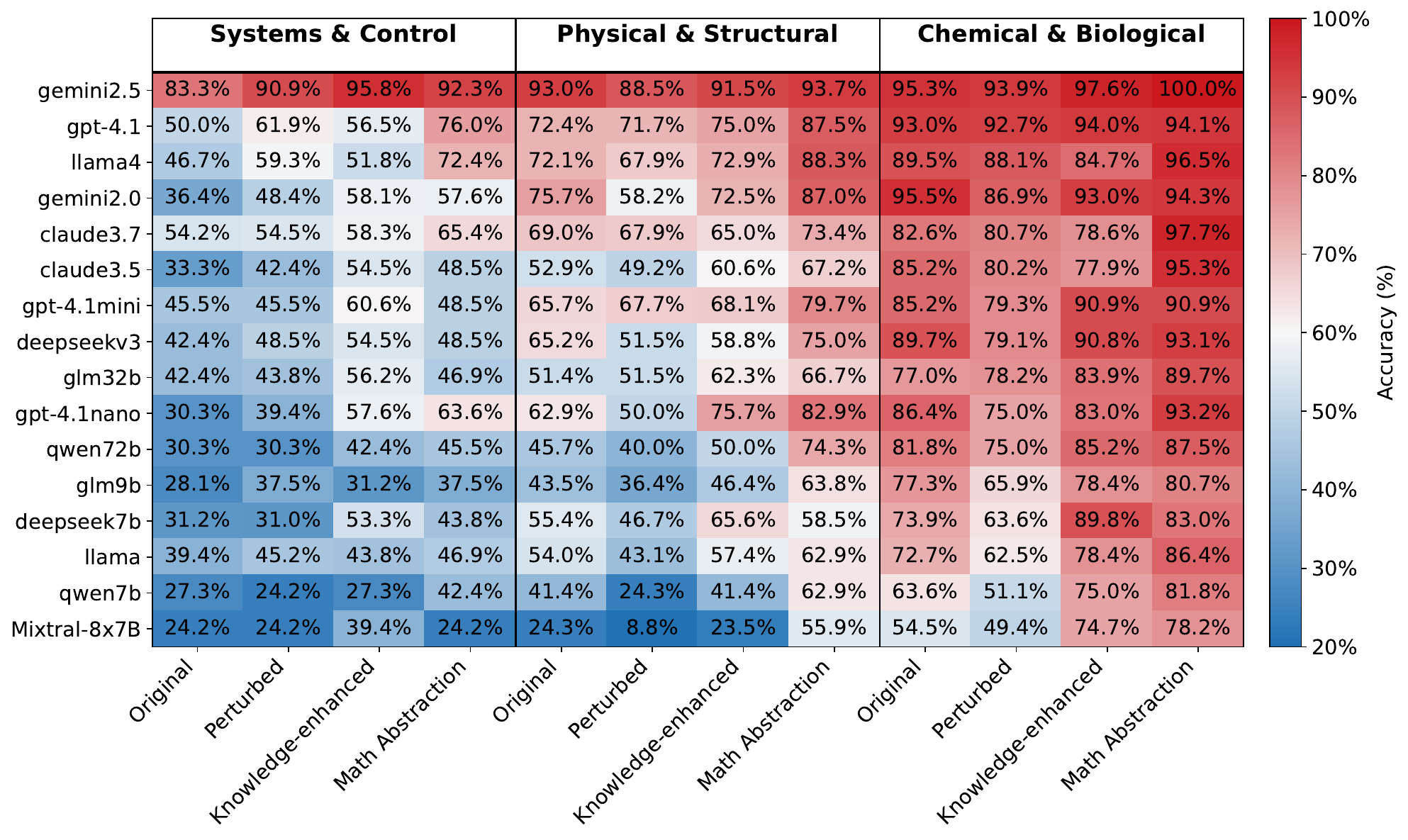} 
    \caption{Accuracy across engineering subfields and problem variants in Level~2.}
    \label{fig:level2_heatmap}
\end{figure*}

Figures~\ref{fig:level1_heatmap} and~\ref{fig:level2_heatmap} present an overview of model accuracy across engineering subfields and problem variants for Level~1 and Level~2, respectively.

\textbf{Model performance varies substantially across engineering subfields.}\quad 
Chemical and biological engineering demonstrates the strongest robustness, with large models maintaining accuracies above 85\%, while structural and physical engineering achieves 70--80\% and systems and control engineering performs the worst, with large models dropping to 60--70\% and small models often below 40\%. These results suggest that robustness to contextual perturbations is closely tied to the task characteristics: chemical and biological problems rely more on formulaic knowledge and are less sensitive to input variations, whereas systems and control problems involve more complex reasoning chains and are more vulnerable to perturbations.

\textbf{Problem variants reveal subfield-specific differences in knowledge use, reasoning, and robustness, showing that these abilities differ significantly between engineering domains.}\quad 
The knowledge-enhanced variant substantially improves performance in chemical and biological engineering, moderately benefits structural and physical engineering, and shows limited gains in systems and control engineering, suggesting the latter’s inability to effectively leverage explicit knowledge. Similarly, the math abstraction variant, which isolates mathematical reasoning by removing context, favors chemical and biological engineering, followed by structural and physical engineering, while systems and control engineering remains the weakest. These patterns indicate that the ability to utilize injected knowledge and maintain mathematical reasoning varies considerably across subfields.

\textbf{The robustness and capability differences across subfields become even more evident under higher task complexity in Level~2.}\quad
Compared to Level~1, Level~2 shows larger performance drops under perturbed inputs, highlighting more severe robustness issues. The positive effects of knowledge-enhanced and math abstraction variants remain concentrated in chemical and biological engineering, with only marginal improvements in structural and physical engineering and negligible gains in systems and control engineering. This indicates that in more complex reasoning and contextual integration tasks, current large language models struggle even more to handle input perturbations, exploit external knowledge effectively, and maintain consistent reasoning, further widening the capability gap across subfields.

\end{document}